\documentclass{article}
\usepackage[submission]{colm2026_conference}

\usepackage{hyperref}
\usepackage{url}
\usepackage{booktabs}
\usepackage{lineno}
\usepackage{color}
\usepackage{tcolorbox}
\usepackage{caption}
\tcbuselibrary{breakable, skins}

\newtcolorbox{promptbox}[1]{
    breakable,
    colback=white,
    colframe=black!70,
    fonttitle=\scriptsize\bfseries,
    title=#1,
    toprule=1.5pt,
    bottomrule=0.5pt,
    left=3pt, right=3pt, top=3pt, bottom=3pt,
}
\newtcolorbox{systembox}{
    breakable,
    colback=white,
    colframe=black!40,
    leftrule=3pt,
    rightrule=0.4pt,
    toprule=0.4pt,
    bottomrule=0.4pt,
    fontupper=\scriptsize,
    title={\scriptsize\textbf{System}},
    left=3pt, right=3pt, top=2pt, bottom=2pt,
    before skip=2pt, after skip=2pt,
}
\newtcolorbox{userbox}{
    breakable,
    colback=white,
    colframe=black!20,
    leftrule=3pt,
    rightrule=0.4pt,
    toprule=0.4pt,
    bottomrule=0.4pt,
    fontupper=\scriptsize,
    title={\scriptsize\textbf{User}},
    left=3pt, right=3pt, top=2pt, bottom=2pt,
    before skip=2pt, after skip=2pt,
}

\definecolor{ForestGreen}{rgb}{0.133,0.545,0.133}
\definecolor{BrickRed}{rgb}{0.796,0.255,0.329}
\definecolor{darkblue}{rgb}{0,0,0.5}
\definecolor{mygreen}{rgb}{0.133,0.545,0.133}
\hypersetup{
  colorlinks=true,
  citecolor=darkblue,
  linkcolor=darkblue,
  urlcolor=darkblue
}

\usepackage{amsmath,amsfonts,bm}

\def\eqref#1{equation~\ref{#1}}

\def\1{\bm{1}}

\DeclareMathAlphabet{\mathsfit}{\encodingdefault}{\sfdefault}{m}{sl}
\SetMathAlphabet{\mathsfit}{bold}{\encodingdefault}{\sfdefault}{bx}{n}

\usepackage{xcolor}
\usepackage{graphicx}
\usepackage{tabularx}
\usepackage{amsmath}
\usepackage[capitalize]{cleveref}
\usepackage{mathtools}
\usepackage{multirow}
\usepackage{setspace}
\hypersetup{colorlinks,linkcolor={blue},citecolor={blue},urlcolor={blue}}

\usepackage{latexsym}
\usepackage{comment}
\usepackage{tabulary}
\usepackage{tabularx}
\usepackage{amsmath}
\usepackage{amsfonts} 
\usepackage{array}

\usepackage[T1]{fontenc}
\usepackage[utf8]{inputenc}
\usepackage{listings}
\usepackage{pifont}
\usepackage{graphicx}
\usepackage{inconsolata}

\usepackage{array}
\usepackage{graphicx}
\usepackage{tabulary}
\usepackage{colortbl}
\usepackage{booktabs}
\usepackage{enumitem}
\usepackage{makecell}
\usepackage{comment}

\usepackage{tcolorbox}   
\usepackage{subcaption} 
\usepackage{wrapfig}
\usepackage{float}

\newif\ifshowcomments
\showcommentsfalse

\newcommand{\algname}{\texttt{RetailSim}}

\definecolor{absgray}{RGB}{242,243,245}
\definecolor{metablue}{RGB}{0,102,204}

\newcommand{\customabstractpage}{
\begin{tcolorbox}[
    enhanced,
    colback=absgray,
    colframe=absgray,
    boxrule=0pt,
    arc=8pt,
    left=3mm,
    right=3mm,
    top=3mm,
    bottom=3mm
]

{\Large\bfseries
What Makes a Sale? Simulating End-to-End Seller--Buyer Retail Dynamics with LLM Agents
\par}

\vspace{3mm}

\textbf{Jeonghwan Choi}$^{1}$, \textbf{Jibin Hwang}$^{1}$, \textbf{Gyeonghun Sun}$^{1}$, \textbf{Minjeong Ban}$^{1}$,\\ \textbf{Taewon Yun}$^{1}$, \textbf{Hyeonjae Cheon}$^{2}$, \textbf{Hwanjun Song}$^{1,3,*}$\par

\vspace{1mm}

$^{1}$Korea Advanced Institute of Science and Technology (KAIST)\\
$^{2}$GS Retail, $^{3}$Cluvion\par

\vspace{1mm}

\texttt{\{hwani.choi, songhwanjun\}@kaist.ac.kr}\par

\vspace{4mm}

\noindent
Evaluating retail strategies before deployment is difficult, as outcomes are determined across multiple stages, from seller-side persuasion through buyer–seller interaction to purchase decisions. However, existing retail simulators capture only partial aspects of this process and do not model cross-stage dependencies, making it difficult to assess how early decisions affect downstream outcomes. We present \algname{}, an end-to-end retail simulation framework that models this pipeline in a unified environment, explicitly designed for simulation fidelity through diverse product spaces, persona-driven agents, and multi-turn interactions. We evaluate \algname{} with a dual protocol comprising human evaluation of behavioral fidelity and meta-evaluation against real-world economic regularities, showing that it successfully reproduces key patterns such as demographic purchasing behavior, the price–demand relationship, and heterogeneous price elasticity. We further demonstrate its practical utility via decision-oriented use cases, including persona inference, seller–buyer interaction analysis, and sales strategy evaluation, showing \algname{}'s potential as a controlled testbed for exploring retail strategies. 

\vspace{4mm}

\noindent
\begin{minipage}[t]{0.65\textwidth}
{\small
\textbf{Date:} July 21, 2026 \par
\textbf{Correspondence:} Hwanjun Song at {\color{metablue}\href{mailto:songhwanjun@kaist.ac.kr}{songhwanjun@kaist.ac.kr}} \par
\textbf{First Author:} Jeonghwan Choi at {\color{metablue}\href{mailto:hwani.choi@kaist.ac.kr}{hwani.choi@kaist.ac.kr}} \par
\textbf{Simulation Demo:} {\color{metablue} \url{https://disl-lab.github.io/RetailSim}}
}
\end{minipage}
\hfill
\begin{minipage}[]{0.27\textwidth}
\vspace*{-0.3cm}
\raggedleft
\includegraphics[width=1.0\linewidth]{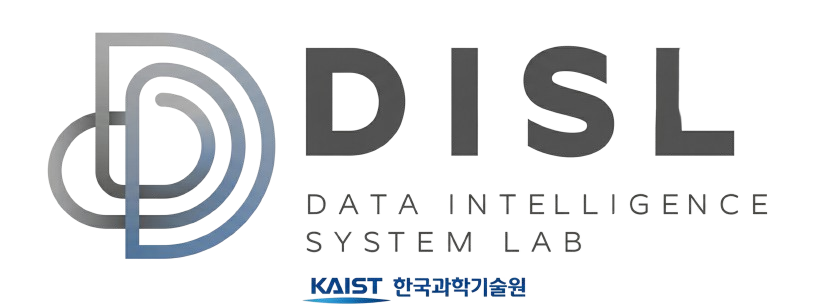}
\end{minipage}

\end{tcolorbox}
}

\begin{document}

\lhead{Published as a conference paper at COLM 2026}
\thispagestyle{empty}

\vspace*{-1.3cm}
\customabstractpage

\section{Introduction}
\label{sec:introduction}


Designing effective retail strategies requires understanding how decisions made at one stage influence downstream outcomes across interdependent seller--buyer interactions \citep{ahearne2022future, wang2025ecombench, zhou2026mixecom}, particularly in terms of purchase conversion, revenue, and post-purchase feedback. However, this remains a long-standing challenge due to two fundamental sources of complexity. First, retail environments involve multiple interacting parties with diverse preferences and behaviors \citep{shi2025you, li2025consumer}, leading to heterogeneous and often unpredictable interaction dynamics. Second, retail interactions unfold over multiple stages \citep{mukherjee2021webrooming, yao2025taubench, gupta2026customer}, such as sales pitch, pre-purchase inquiry handling, purchase decision, and post-purchase support, making it difficult to trace how decisions propagate across stages and ultimately determine final retail outcomes.

To address these challenges, simulation offers a promising approach by enabling controlled and repeatable analysis of complex interactions without the cost and irreversibility of real-world deployment \citep{sturley2018evaluating, jamali2026agent}. Recent advances in large language models (LLMs) further enhance this paradigm by enabling realistic and scalable multi-agent simulations, opening up new opportunities for studying human-like behaviors under controlled interaction scenarios \citep{zhang2025simulating, chu2025simul1, hu2026simbench}. In the retail domain, such environments allow systematic variation of strategies, participant characteristics, and interaction conditions, making it possible to directly measure their impact on key outcomes, including customer service success \citep{yao2025taubench, zhou2026mixecom}, recommendation quality \citep{wang2025ecomscriptbench, wang2025shoppingbench}, and negotiation performance \citep{xia2024bargaining, bianchi2024negotiationarena, zhu2025automatedrisky}.

\begin{figure}[t]
\centering
\hspace*{-0.3cm}
\includegraphics[width=14.4cm]{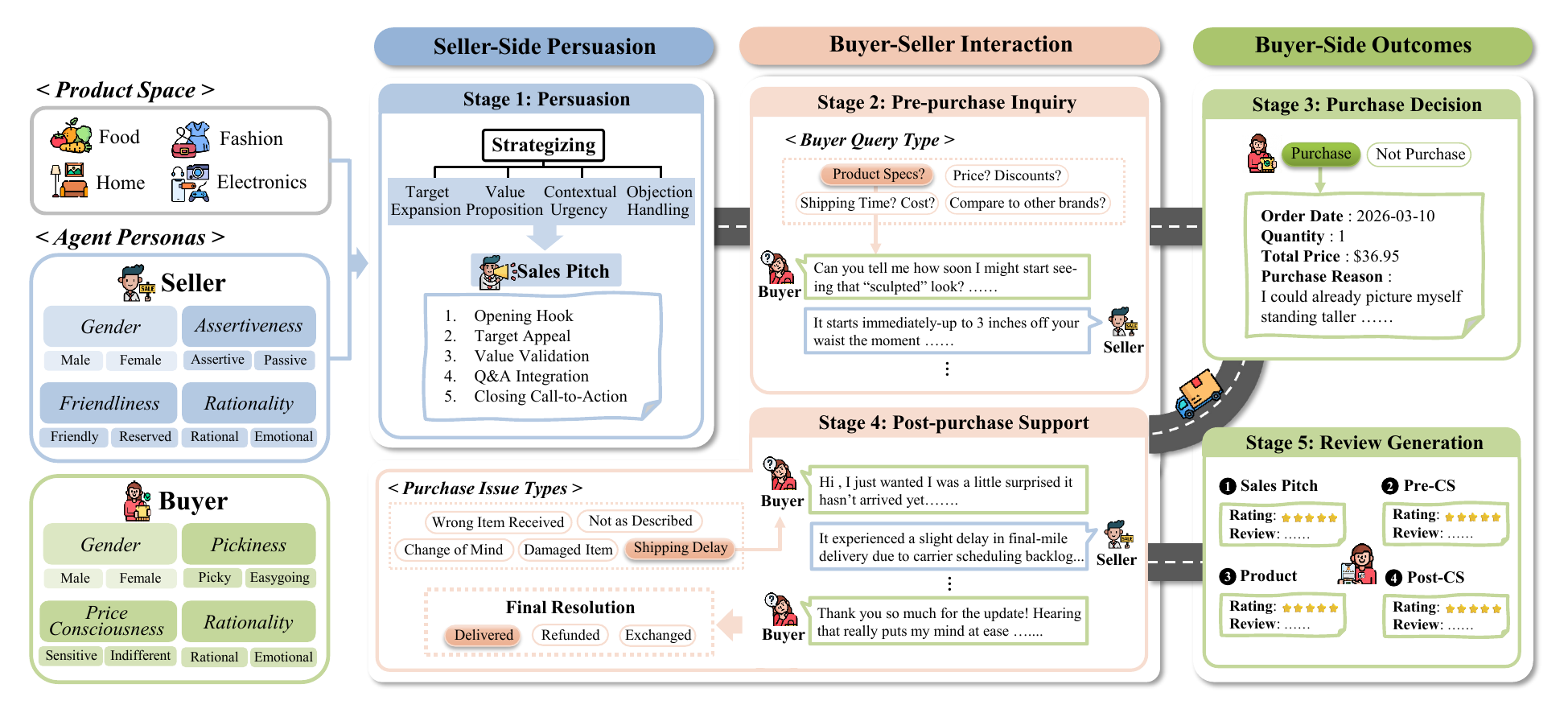}
\vspace*{-0.9cm}
\caption{Overview of \algname{}: The framework models retail interactions as a unified, multi-stage pipeline, capturing seller strategies, multi-turn interactions, and downstream buyer outcomes, enabling end-to-end analysis of how decisions propagate across stages.}
\label{fig:overview}
\vspace*{-0.55cm}
\end{figure} 

Despite these advances, existing studies \citep{wang2025shoppingbench, zhu2025automatedrisky, zhou2026mixecom} primarily focus on specific components of the retail process, such as shopping, negotiation, or customer support, without modeling the full progression from seller-side persuasion to buyer-side outcomes within a unified framework. As a result, they do not capture how early decisions influence downstream outcomes, making it difficult to verify whether resulting behaviors remain consistent with fundamental economic principles (\emph{e.g.}, price–demand relationships). Consequently, the economic consistency of such simulations remains unclear, raising concerns about their reliability for decision support \citep{wang2025opera, mansour2025paars, bansal2025magenticmarketplace}. These limitations motivate a unified framework for end-to-end analysis grounded in real-world economic regularities.

To remedy this, we introduce \algname{}, a novel multi-agent simulation framework that models the full retail process in a unified setting, capturing dependencies across interactions and enabling the reproduction of real-world economic regularities. As shown in Figure \ref{fig:overview}, it simulates the real-world retail interaction process  as it unfolds over time, including "seller--side persuasion" (\emph{i.e.}, sales pitch generation), "buyer--seller interactions" (\emph{i.e.}, pre- and post-purchase inquiries), and "buyer-side outcomes" (\emph{i.e.}, purchase decisions and review generation), partially treated in prior work. The framework further incorporates three core components to capture real-world complexity, including (i) diverse product categories, (ii) seller and buyer personas capturing behavioral traits, and (iii) multi-turn interactions before and after purchase, covering scenarios such as inquiries, complaints, returns, and refunds. All together, the unified multi-stage design and these components enable controlled, realistic variation and detailed analysis of interaction outcomes.


Beyond the framework, we establish an empirical evaluation protocol that jointly assesses simulation fidelity and its effectiveness. First, for \emph{fidelity}, we propose a two-level, complementary simulation evaluation: at the {stage} level, we perform human evaluation to verify whether generated interactions align with intended personas and stage-specific behaviors; and at the {system} level, we go beyond behavioral fidelity and test whether the simulation preserves {economic consistency} by reproducing fundamental real-world regularities (\emph{e.g.}, the negative relationship btw price and purchase). Second, for \emph{effectiveness}, we demonstrate that \algname{} enables decision-oriented analysis by supporting controlled interventions and revealing their downstream effects across the full interaction pipeline. Specifically, we showcase three applications: inferring latent personas from interaction data, analyzing persona-conditioned revenue patterns, and identifying effective sales strategies for improving conversion and revenue. Our work differs from prior studies in four key aspects:

\noindent(1) We introduce \algname{}, a causally structured, end-to-end simulation framework that explicitly models cross-stage dependencies in retail interactions, enabling analysis of how early-stage decisions propagate to downstream economic outcomes.

\noindent(2) We develop a controllable simulation environment with persona-driven agents, where behavioral factors, including price sensitivity and rationality, can be systematically varied to study their downstream impact on interaction dynamics and outcomes.

\noindent(3) We propose a system-level evaluation paradigm based on economic consistency, introducing meta-evaluation against real-world behavioral regularities as a criterion for assessing LLM-based simulations beyond surface-level fidelity.

\noindent(4) We demonstrate that \algname{} serves as a scientific testbed for analyzing LLMs as economic agents, enabling measurement of latent behavioral traits and their interaction effects on revenue and decision outcomes.

\vspace*{-0.15cm}
\section{Related Work}
\label{sec:related-work}
\vspace*{-0.15cm}

\textbf{Simulation for Complex Interaction Modeling.}
Modeling multi-stage interactions is challenging as outcomes arise from interdependent decisions over time and are difficult to observe or control in real-world settings, such as healthcare \citep{almansoori2025self} and retail \citep{wang2025ecombench}. Simulation mitigates this by enabling controlled, repeatable analysis of how decisions propagate across stages \citep{jamali2026agent}. Recent advances in LLMs further extend this paradigm, enabling realistic, multi-turn, and grounded interaction simulation, with growing applications in retail and economic domains \citep{zhang2025simulating, chu2025simul1, hu2026simbench}. Building on this, prior retail-focused work largely targets individual stages: post-purchase service \citep{wang2025ecombench, zhou2026mixecom}, recommendation and intent matching \citep{wang2025ecomscriptbench, wang2025shoppingbench, zhao2026ecomstage}, and buyer–seller negotiation \citep{xia2024bargaining, bianchi2024negotiationarena, zhu2025automatedrisky}. Persona-based simulators further introduce heterogeneous agent behaviors to better capture buyer decision styles \citep{wang2025customerr1, bansal2025magenticmarketplace, zhang2026shopr1}. 

\textbf{Economic and Behavioral Foundations of Retail.}
Evaluating retail strategies requires grounding in behavioral and economic principles, as retail outcomes emerge from the accumulation of seller communication, buyer responses, and post-purchase engagement across stages rather than isolated interactions \citep{wang2025ecombench, zhou2026mixecom}. In this regard, prior work highlights the importance of personalization and interaction context in shaping conversion and satisfaction beyond static matching \citep{wang2025ecomscriptbench, zhao2026ecomstage}, and recent LLM-based simulations suggest such complex socio-economic behaviors can be reproduced at scale \citep{li2024econagent, chu2025simul1, feng2025simcity}. 

However, prior studies primarily establish behavioral fidelity at the interaction level, with limited consideration of how seller–buyer interactions connect across the full process or whether their aggregate outcomes reflect real-world economic regularities \citep{yao2025taubench, wang2025ecombench, zhu2025automatedrisky, wang2025shoppingbench, bansal2025magenticmarketplace, zhou2026mixecom}.
In addition, they focus on optimizing task-specific outcomes, such as customer service success, recommendation quality, and negotiation performance, without considering how decisions propagate across interactions. As a result, their applicability to real-world use cases such as strategy evaluation and decision support remains limited. We address this gap with a unified framework for joint analysis of seller--buyer dynamics and economic outcomes. Table \ref{tab:comparison} highlights how our approach differs from 13 existing retail simulation frameworks across key dimensions, including persona-driven role modeling and stage coverage.

\vspace*{-0.15cm}
\section{RetailSim Framework}
\label{sec:method}
\vspace*{-0.15cm}



We introduce a cross-stage dependency model for end-to-end retail simulation, linking seller-side persuasion, buyer–seller interaction, and buyer-side outcomes through structured state transitions. This enables both realistic generation and causal analysis of downstream outcomes. We instantiate this formalization with three key components for simulation fidelity: diverse product spaces, persona-driven agents, and multi-turn interactions.

\vspace*{-0.1cm}
\subsection{Unified Multi-Stage Interaction Simulation}
\label{sec:method_framework}
\vspace*{-0.1cm}

Retail interactions naturally follow a multi-stage process spanning seller-side persuasion, buyer–seller interaction, and buyer-side outcomes, as well established in customer journey modeling \citep{mukherjee2021webrooming, gupta2026customer}. To reflect this real-world process, \algname{} instantiates a causally structured, end-to-end simulation pipeline across the three stages, where each stage passes its outputs forward as context and interacts with prior stages. Prompts for agent behaviors are provided in Tables \ref{tab:seller_persona}--\ref{tab:review_4} in Appendix \ref{appendix:llm_prompt}.

\textbf{Seller-side Persuasion.} 
The simulation process begins with seller-side strategy formulation and sales pitch generation to drive product sales. Given a product, the seller agent generates a sales pitch, a "product-specific sales script" (see an example in Table 
\ref{tab:simulation_output_stage1_strategy}--\ref{tab:simulation_output_stage1_pitch}), by prompting an LLM to create the pitch along the following four key dimensions.
\begin{itemize}[leftmargin=*, noitemsep, topsep=0pt]
\item \emph{Target Expansion}: Defining core and secondary target audiences to broaden the reach of the pitch beyond the most direct buyers.
\item \emph{Value Proposition}: Translating product features into buyer benefits and justifying price--quality trade-offs via competitive positioning, technical credibility, or cost-per-use logic.
\item \emph{Contextual Urgency}: Establishing a product-specific reason to buy now, grounded in situational or seasonal milestones relevant to the buyer.
\item \emph{Objection Handling}: Anticipating likely buyer hesitations and preparing persuasive responses to turn concerns into selling points.
\end{itemize}
By structuring persuasion in this way, the framework generates product-aware, practically grounded sales pitches\footnote{The key dimensions are derived from analysis of seven high-quality real-world sales scripts provided by a retail company with over 5,000 employees. See the details in Appendix \ref{appendix:sales_strategy}.} that reflect real retail strategies and provide a consistent starting point for subsequent buyer--seller interactions.

\textbf{Buyer--Seller Interactions.} Following the generated sales pitch, the buyer is presented with a product-specific sales script, after which interactions proceed around issues arising before and after purchase (\emph{e.g.}, clarifications, concerns, delivery, refund). This process is critical, as outcomes are shaped not only by the initial pitch but by how uncertainties and issues are addressed throughout the buyer–seller interaction \citep{edelman2015competing, lemon2016understanding}. We model this process as "five-turn buyer–seller conversations": before purchase, interactions capture decision-related issues, and after purchase, they continue for buyers who complete a transaction. All issue types are summarized in Table \ref{tab:interaction_scenarios}.
\begin{itemize}[leftmargin=*, noitemsep, topsep=0pt]
\item \emph{Pre-inquiry}: The buyer selects inquiry topics of interest in \{product specifications, product comparisons, price/discount, shipping\} and engages in conversation with the seller to resolve uncertainties before deciding whether to purchase.
\item \emph{Post-inquiry}: Only if the buyer completes a purchase, they are assigned an issue type in \{shipping delay, wrong item received, change of mind, damaged, not as described\} and interact with the seller to reach an outcome in \{delivered, exchanged, refunded\}.
\end{itemize}

\textbf{Buyer-side Outcomes.} Buyer-side outcomes occur at two points in the simulation process. After the pre-purchase inquiry stage, the buyer agent makes a "purchase decision," and after post-purchase interactions, the buyer generates "product reviews."
\begin{itemize}[leftmargin=*, noitemsep, topsep=0pt]
\item \emph{Purchase Decision}: The buyer agent decides whether to purchase and how many units to buy, along with a rationale, conditioned on the accumulated context of the sales pitch (product-specific script) and pre-purchase inquiries.
\item \emph{Product Review}: Only for buyers who complete a purchase, the buyer generates four-dimensional reviews based on the full history, covering sales pitch effectiveness, pre-purchase support quality, post-purchase resolution, and product satisfaction.
\end{itemize}

Together, this framework captures the full retail process, enabling end-to-end analysis of how interactions across stages determine final outcomes.

\subsection{Key Components for Simulation Fidelity}
\label{sec:method_preliminary}

While the unified pipeline enables end-to-end simulation, capturing realistic behavior requires more than modeling the interaction flow. To this end, \algname{}~ incorporates key components that capture real-world variation in products, agent behaviors, and interaction scenarios. These components enable realistic, controlled simulation of retail interactions and systematic analysis of how product, persona, and interaction factors drive outcomes.

\textbf{Product Space.} A realistic product space is essential for capturing variation in buyer preferences and decision contexts \citep{ding2023impact}. Accordingly, we construct a realistic product space using publicly available product metadata from the Amazon Reviews`23 dataset \citep{hou2024bridging}. During simulation, we randomly sample a fixed number of products covering $34$ categories, which are grouped into four high-level categories of \{Food, Fashion, Home, Electronics\} for ease of exposition\footnote{The products are categorized into the four high-level product categories using Qwen3-235B-A22B-Instruct based on their original title, description, and price.}. For all simulation experiments, we sample an equal number of products from each category to ensure balanced coverage.

\textbf{Agent Personas.} Agent personas capture heterogeneity in seller behaviors and buyer preferences throughout the end-to-end process, influencing interaction dynamics and outcomes \citep{gao2024large, mansour2025paars}. In our framework, we consider two agent roles, "sellers" and "buyers," and assign each agent a demographic attribute (gender, male or female) along with three role-specific behavioral traits for each role, which are injected into the LLM via persona-specific prompts in Appendix \ref{appendix:llm_prompt}. This design enables controlled variation of seller and buyer personas, supporting diverse simulation scenarios and systematic analysis of their impact on interaction dynamics and outcomes.

Specifically, the seller agent is characterized by \emph{assertiveness} (assertive vs. passive), which controls the intensity and directness of persuasion; \emph{friendliness} (friendly vs. reserved), which governs interpersonal tone from casual and relational to formal and business-like; and \emph{rationality} (rational vs. emotional), which shapes persuasive content from evidence-based arguments to affective and narrative appeals. In contrast, the buyer agent is characterized by \emph{pickiness} (picky vs. easygoing), which determines the strictness of evaluation over product quality, performance, and details; \emph{price consciousness} (sensitive vs. indifferent), which governs how strongly price and value-for-money are weighted in purchase decisions; and \emph{rationality} (rational vs. emotional), which shapes whether decisions are driven by factual evaluation or intuitive impressions. For both roles, each trait is defined as a choice between pre-defined values for tractable analysis across multi-dimensional traits.

\textbf{Interaction Scenarios.} Additionally, interaction scenarios capture the diversity of real-world buyer--seller exchanges and context-dependent dynamics that influence outcomes \citep{lemon2016understanding, wang2025ecom}. We thus model realistic scenarios covering issues before and after purchase, including product inquiries, comparisons, pricing and shipping questions, as well as post-purchase issues such as delivery problems, returns, and refunds (as summarized in Table \ref{tab:interaction_scenarios} and \ref{tab:return_refund_policy}). These interactions are realized as short dialogues, where each response is conditioned on the accumulated simulation trajectory up to that point to ensure factual consistency and coherent progression of the interaction.

\vspace*{-0.1cm}
\section{Simulation Fidelity and Economic Consistency}
\label{sec:evaluation}
\vspace*{-0.1cm}

We evaluate simulation fidelity at two complementary levels: at the \emph{stage} level, we assess whether agents faithfully follow the stage-specific behaviors (\emph{e.g.}, persuasive strategy execution, inquiry handling) and provided personas; at the \emph{system} level, we examine whether the resulting outcomes align with real-world economic principles. These evaluations validate the reliability of \algname{} as a simulation testbed, which the decision-oriented analyses in Section \ref{sec:application} build upon.

For the backbone model, \algname{} can be instantiated with any LLM. We thus evaluate eight LLMs under a homogeneous setup using the same LLM for both seller and buyer agents, enabling the comparison of their simulation capabilities. The eight LLMs include both open- and closed-source models of varying sizes: Qwen3-Next-80B-A3B-Instruct, Qwen3-235B-A22B-Instruct, GPT-oss-120B, DeepSeek-V3.2, Gemini-3-Flash, Gemini-3.1-Pro, GPT-5.4-mini, and GPT-5.4. Heterogeneous setups are discussed in {Section \ref{sec:application}}.

\vspace*{-0.05cm}
\subsection{Stage Level Evaluation via Human Evaluation}
\label{sec:stage-level-eval}
\vspace*{-0.05cm}

At the stage level, we run simulations with \algname{} and validate their fidelity through human evaluation, following prior work \citep{majumder2020hiking, hwang2026personality}. We evaluate two complementary aspects: (i) we assess \emph{task fidelity} across five components—sales scripts, pre-purchase inquiries, purchase decisions, post-purchase inquiries, and product reviews—by measuring whether generated outputs adhere to the simulation guidelines and exhibit realistic, coherent behavior, using a 5-point Likert scale (Very Poor to Excellent); and (ii) we evaluate \emph{persona fidelity} by varying persona traits and testing whether they are reflected in agent behavior. Specifically, human annotators are shown A/B pairs of outputs generated under opposing traits (\emph{e.g.}, rational vs. emotional in rationality) and asked to choose which output reflects the target trait (\emph{e.g.}, rational); accuracy (Acc) is measured by whether annotators select the correct one.

For this evaluation, we run simulations on 12 products across all combinations of persona traits for both seller and buyer agents, using each LLM backbone. For task fidelity, we sample $3$ outputs per task for each product, resulting in a total of $8$(LLMs) $\times$ $12$(products) $\times$ $5$(tasks) $\times$ $3$(samples) $=$ 1,440 outputs for human evaluation. For persona fidelity, we construct A/B pairs by comparing outputs generated under opposing conditions for a target trait, while keeping the other traits fixed. Thus, for each target trait, we collect four A/B pairs, as there are four possible combinations of the remaining two traits. In total, this results in $8$(LLMs) $\times$ $12$(products) $\times$ $2$(roles) $\times$ $3$(target traits) $\times$ $4$(combinations of remaining traits) $=$ 2,304 A/B pairs for human evaluation. Each task output and each A/B pair is annotated by three human annotators via Amazon MTurk, and we report inter-annotator agreement (IAA) using Krippendorff’s $\alpha$ to assess annotation reliability ($\alpha \geq 0.67$: meaningful reliability \citep{cook2025efficient, ban2026completing}). Details are in Appendix \ref{appendix:human_annotation}.

\begin{table*}[t]
\centering
\scriptsize
\vspace{-0.4cm}
\setlength{\tabcolsep}{2.7pt}
\begin{tabular}{l|l|l|cccccccc|c}
\toprule
Aspect & \multicolumn{2}{c|}{Category}
  & \makecell{Qwen3\\-80B} 
  & \makecell{Qwen3\\-235B} 
  & \makecell{GPT-oss\\-120B} 
  & \makecell{DeepSeek\\-V3.2} 
  & \makecell{Gemini-3\\-Flash} 
  & \makecell{Gemini-3.1\\-Pro} 
  & \makecell{GPT-5.4\\-mini} 
  & \makecell{GPT\\-5.4} 
  & \makecell{Kripp.'s\\$\alpha$ (IAA)} \\
\midrule
\multirow{5}{*}{~~Task}
  & \multicolumn{2}{c|}{Sales Script}       & 4.36 & 4.50 & 4.45 & 4.29 & 4.28 & 4.40 & 4.30 & 4.51 & 0.67 \\
  & \multicolumn{2}{c|}{Pre-purchase Inquiry}     & 2.73 & 3.56 & 3.20 & 3.32 & 3.10 & 2.72 & 4.45 & 4.20 & 0.77 \\
  & \multicolumn{2}{c|}{Purchase Decision}        & 3.21 & 3.92 & 3.67 & 3.76 & 3.94 & 3.96 & 3.99 & 4.10 & 0.71 \\
  & \multicolumn{2}{c|}{Post-purchase Inquiry}    & 2.99 & 3.65 & 3.75 & 3.86 & 3.41 & 2.90 & 3.14 & 4.06 & 0.77 \\
  & \multicolumn{2}{c|}{Product Review}           & 4.32 & 4.27 & 3.73 & 3.99 & 3.94 & 3.97 & 4.14 & 4.13 & 0.71 \\\midrule
\multirow{6}{*}{\!\!Persona}
  & \multirow{3}{*}{Seller} & Assertiveness     & 99.7 & 99.3 & 97.9 & 97.2 & 99.7 & 99.3 & 91.7 & 95.0 & 0.81 \\
  &                         & Friendliness      & 99.3 & 96.5 & 99.7 & 99.3 & 99.3 & 99.3 & 94.3 & 93.6 & 0.83 \\
  &                         & Rationality       & 90.3 & 84.0 & 88.7 & 91.0 & 97.9 & 99.7 & 84.8 & 86.8 & 0.74 \\
\cmidrule{2-12}
  & \multirow{3}{*}{Buyer}  & Pickiness         & 98.2 & 99.4 & 97.3 & 98.2 & 99.1 & 99.7 & 91.0 & 97.6 & 0.81 \\
  &                         & Price Consciousness & 97.6 & 97.6 & 99.4 & 99.1 & 99.4 & 99.7 & 93.4 & 99.4 & 0.77 \\
  &                         & Rationality       & 99.7 & 99.4 & 97.6 & 99.7 & 99.7 & 99.7 & 97.0 & 99.7 & 0.83 \\
\bottomrule
\end{tabular}
\vspace{-0.2cm}
\caption{Simulation fidelity of \algname{} with respect to task fidelity (1--5 Likert-scale evaluation) and persona fidelity (accuracy (\%) via A/B comparison based evaluation).}
\vspace{-0.4cm}
\label{tab:annotation_result}
\end{table*}


\noindent\textbf{Results.} Table \ref{tab:annotation_result} summarizes human evaluation results in terms of task and persona fidelity. Overall, \algname{} achieves high task and persona fidelity across all LLM backbones. The Krippendorff $\alpha$ ranges from 0.67 to 0.83, supporting the reliability of our evaluation. In detail, for task fidelity, the sales script and product review tasks consistently achieve strong performance across all LLMs, with scores generally exceeding 4 out of 5. In contrast, for pre-/post-purchase inquiry and purchase decision tasks, larger or proprietary LLMs tend to achieve higher scores, suggesting that these tasks require more advanced reasoning and interaction capabilities. Contrarily, for persona fidelity, all LLMs achieve consistently high accuracy across both roles, with most scores exceeding 95\%. This suggests that persona attributes are generally preserved across LLMs. This suggests that \algname{} produces a largely consistent persona-conditioned behavior in simulation.

\vspace{-0.1cm}
\subsection{System Level Evaluation via Meta Evaluation}
\label{sec:meta_evaluation}
\vspace{-0.1cm}

A simulation can appear realistic at each stage yet fail to reproduce real-world behavioral and economic patterns. However, this system-level validity has not been explicitly verified in prior work. Therefore, we propose system-level "{meta evaluation}," where we test whether the simulation reproduces well-established, empirically observed patterns in real-world retail. We validate three economic aspects: (i) \emph{demographic purchasing patterns}; (ii) \emph{price–demand relationship}; and (iii) \emph{heterogeneous price elasticity}. We run simulation using \algname{} but with slightly different setups for each of the three aspects. Beyond these, we further disentangle framework-induced patterns from the models' emergent behavior, and validate simulated review ratings against real-world rating data.

\textbf{Demographic Purchasing Pattern.} We examine whether \algname{} captures a representative pattern in retail, namely gender-dependent differences in purchasing behavior \citep{kanwal2021gender}. To this end, we run simulations in the "Fashion" category using 15 products (5 male-oriented, 5 female-oriented, and 5 unisex). The same setup is applied across all eight LLM backbones to avoid model-specific bias, resulting in $8$(LLMs) $\times$ $15$(products) 
\begin{wraptable}{r}{0.4\textwidth}
\centering
\scriptsize
\setlength{\tabcolsep}{4.4pt}
\vspace{-0.3cm}
\begin{tabular}{lccc}
\toprule
Product & Male & Female & $\Delta$(M$-$F) \\
\midrule
Men-oriented   & 54.6\% & 48.3\% & $+$6.3\%p$^{*}$  \\
Women-oriented & 32.5\% & 55.8\% & $-$23.3\%p$^{*}$ \\
Unisex         & 60.8\% & 62.5\% & $-$1.7\%p         \\
\bottomrule
\end{tabular}
\vspace{-0.25cm}
\caption{Purchase rates by buyer gender across product types, with differences ($\Delta$), where *: $p < 0.05$.}
\label{tab:meta_gender}
\vspace{-0.25cm}
\end{wraptable}
$\times$ $16$(buyer persona combinations, including gender) $=$ 1,920 simulation runs, with seller personas randomly assigned. Table \ref{tab:meta_gender} reveals that male buyers exhibit higher purchase rates for men-oriented products, while female buyers show substantially higher purchase rates for women-oriented products. In contrast, the difference is negligible for unisex products, indicating that \algname{} successfully captures gender-aligned purchasing patterns.

\begin{wraptable}{r}{0.4\textwidth}
\centering
\scriptsize
\vspace{-0.4cm}
\setlength{\tabcolsep}{1.9pt}
\begin{tabular}{lccccc}
\toprule
& \multicolumn{5}{c}{Price Condition} \\
\cmidrule(lr){2-6}
& $-$10\% & $-$5\% & 0\% & $+$5\% & $+$10\% \\
\midrule
Purchase Rate & 64.8\% & 61.7\% & 57.7\% & 57.3\% & 56.6\% \\
\bottomrule
\end{tabular}
\vspace{-0.25cm}
\caption{Buyers' purchase rates under different price conditions.}
\label{tab:meta_economics}
\vspace{-0.25cm}
\end{wraptable}
\textbf{Price Demand Relationship.} It is also well established in economics that customers' demand decreases as prices increase, and conversely increases as prices decline \citep{hitsch2021prices}. However, it is unclear whether such price–demand dynamics emerge in LLM-driven agent simulations. To verify whether this economic principle emerges with \algname{}, we randomly select eight items from the demographic setup and vary their prices across five conditions ($\pm$10\%, $\pm$5\%, and the original price), conducting additional simulation runs. To examine the price–demand relationship, we analyze how buyers' purchase rates change under different pricing conditions. Table \ref{tab:meta_economics} shows that purchase rates decrease monotonically from 64.8\% at a 10\% discount to 56.6\% at a 10\% premium, confirming that \algname{} reproduces the price–demand relationship observed in economics.

\textbf{Heterogeneous Price Elasticity.} Under the same setup as the price–demand analysis, we examine whether \algname{} captures systematic differences in demand responsiveness across buyers with varying levels of price consciousness, reflecting the well-established principle that more price-sensitive consumers exhibit greater demand elasticity \citep{phillips2021pricing}. We partition buyers into two groups based on the price consciousness trait---{sensitive} and {indifferent}---and compute their per-product price elasticity, defined as the percentage change in purchase rate relative to the percentage change in price (see Eq.~\eqref{eq:price_elasticity} of Appendix \ref{appendix:price_elasticity}), quantifying how strongly purchase probability responds to price variation; The higher the elasticity, the more responsive buyers are to price changes. We observe that price-sensitive buyers exhibit much higher elasticity ($2.58$) than price-indifferent buyers ($0.76$), a large difference of $1.82$ that is statistically significant ($p<0.001$), confirming that \algname{} captures real-world heterogeneity in price responsiveness. Appendix \ref{appendix:persona_estimation} further shows that persona-driven differences in price sensitivity align with real-world behavior.

\begin{wraptable}{r}{0.47\textwidth}
\centering
\scriptsize
\setlength{\tabcolsep}{2.6pt}
\vspace{-0.4cm}
\begin{tabular}{lcccc}
\toprule
& \multicolumn{2}{c}{Gender $\Delta$(M$-$F)} & \multicolumn{2}{c}{Elasticity} \\
\cmidrule(lr){2-3}\cmidrule(lr){4-5}
Condition & Men-or. & Women-or. & $|E_d|$ & Ratio \\
\midrule
Original        & $+$6.3\%p$^{*}$ & $-$23.3\%p$^{*}$ & 0.81 & 3.40$\times$$^{*}$ \\
No-Persona      & $-$2.1\%p       & $-$1.3\%p        & 0.89 & 3.74$\times$$^{*}$ \\
No-Anchor-Price & --              & --               & 0.48 & 2.25$\times$$^{*}$ \\
\bottomrule
\end{tabular}
\vspace{-0.25cm}
\caption{Meta-evaluation results under ablated prompts, where *: $p<0.001$.}
\label{tab:meta_ablation}
\vspace{-0.3cm}
\end{wraptable}

\textbf{Disentangling Framework and Model Effects.} While the above patterns may appear intuitive, it remains unclear whether they emerge from the framework's design or simply from the capability of modern LLMs. To disentangle the two, we re-run the demographic and elasticity experiments under two ablated prompts. \emph{No-Persona} removes the buyer persona block and the gender clause from the purchase decision, and \emph{No-Anchor-Price} further neutralizes the price framing. This results in $8$(LLMs) $\times$ $16$(buyer personas) $\times$ $5$(price conditions) $\times$ $15$(products) $=$ 9,600 additional runs (see Appendix \ref{appendix:ablation} for details). Table \ref{tab:meta_ablation} reveals that the gender-aligned pattern collapses to non-significance under No-Persona, indicating that this regularity emerges through the framework's persona injection rather than the models alone. In contrast, the elasticity gap remains significant under No-Anchor-Price, where buyers are given neither persona definitions nor price framing, reflecting a baseline price reasoning intrinsic to the models. These results indicate that \algname{} converts model capabilities into controlled, economically consistent simulation.

\textbf{Review Rating Validation.} Beyond behavioral regularities, we examine whether the simulated review ratings align with real ratings 
\begin{wraptable}{r}{0.44\textwidth}
\centering
\scriptsize
\setlength{\tabcolsep}{3.2pt}
\vspace{-0.4cm}
\begin{tabular}{lcccc}
\toprule
Model & Pearson & Spearman & Mean & $W_1$ \\
\midrule
Real (Amazon`23) & -- & -- & 4.20 & -- \\
\midrule
Qwen3-235B    & $+$0.60 & $+$0.59 & 3.96 & 0.25 \\
DeepSeek-V3.2 & $+$0.60 & $+$0.58 & 3.86 & 0.34 \\
\bottomrule
\end{tabular}
\vspace{-0.25cm}
\caption{Validation of simulated ratings against real ratings, with product-level correlations, mean ratings, and Wasserstein-1 ($W_1$) distance to the real distribution.}
\label{tab:rating_validation}
\vspace{-0.25cm}
\end{wraptable}
in the Amazon Reviews`23 dataset \citep{hou2024bridging}. To this end, we ground each buyer's product experience in the textual content of real reviews, without exposing their star ratings, and simulate reviews over 120 products, resulting in $2$(LLMs) $\times$ $120$(products) $\times$ $32$(buyer personas) $=$ 7,680 simulated reviews. Table \ref{tab:rating_validation} shows that the simulated per-product mean ratings strongly correlate with the real means, and that the simulated distribution reproduces the J-shaped pattern of real reviews with minor deviations in mid-range ratings. These results confirm that the review mechanism of \algname{} faithfully converts product experience into ratings. Details, including full rating distributions, are in Appendix \ref{appendix:rating_validation}.

\vspace*{-0.20cm}
\section{Application of RetailSim across Three Use Cases}
\label{sec:application}
\vspace*{-0.2cm}

While prior work focuses on evaluating simulation fidelity, it remains unclear how such simulations support practical, decision-oriented analysis despite their importance. To address this gap, we design three use cases that operationalize simulation as a tool for analyzing seller and buyer behavior and their impact on retail outcomes. To mimic realistic user populations, we treat five LLMs as heterogeneous agents without explicit persona injection, allowing each model to act according to its inherent tendencies. This setup captures latent behavioral diversity and enables analysis of how such differences propagate through multi-stage interactions to affect retail outcomes. Furthermore, the framework serves as a benchmark for evaluating LLMs as economic agents through their downstream outcomes.

We conduct simulations across 120 products spanning four high-level categories, with five LLMs each serving as both seller and buyer agents. All pairwise combinations of seller–buyer LLMs are considered, including self-interactions, resulting in $120$(products) $\times$ $25$(seller-buyer LLM combinations) = $3{,}000$ simulation runs. See Appendix \ref{appendix:llm_prompt} for details.

\vspace*{-0.15cm}
\subsection{Use Case 1: Latent Persona Estimation}
\label{sec:app_persona}
\vspace*{-0.15cm}

In real-world settings, user traits are not directly observable and must be inferred from interactions. Estimating such latent personas is essential for personalization, as they influence the purchase conversion and revenue. In this use case, we leverage \algname{} to infer latent personas from interaction traces.

\textbf{Persona Classifier Design.} To enable supervised learning of persona traits, we construct a labeled dataset via simulations with explicit persona injection, where ground-truth traits are controlled in Section \ref{sec:stage-level-eval}, yielding $2{,}304$ labeled outputs per task across five tasks in \algname{}. We then train separate binary classifiers for each trait (three for sellers and three for buyers), each distinguishing opposing trait values (\emph{e.g.}, rational vs. emotional). Rather than using all five task outputs, we perform stage-wise feature selection to identify the most informative task for each trait, and train classifiers on the selected representations only. The resulting models achieve strong performance, with test accuracy ranging from $85.5\%$ to $98.5\%$. Further details are provided in Appendix \ref{appendix:persona_estimation}.

\begin{figure}[!t]
\centering
\includegraphics[width=1\textwidth]{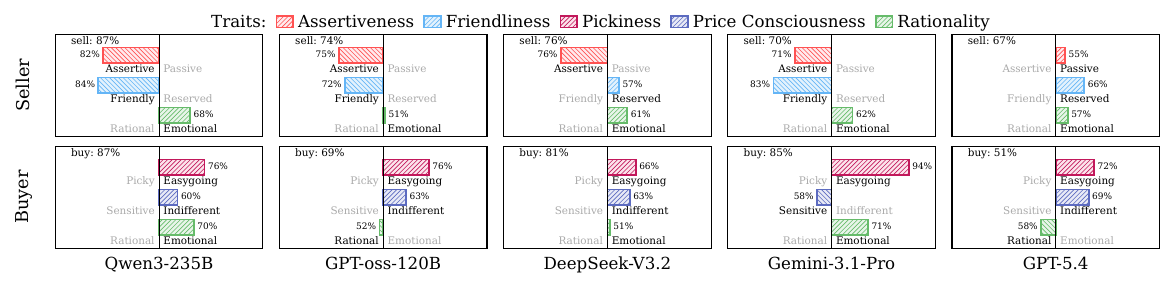}
\vspace{-0.8cm}
\caption{Estimated personas of Five LLMs as seller (top) and buyer (bottom) roles. Each bar represents the probability assigned to the dominant label in a binary trait classification.}
\vspace{-0.2cm}
\label{fig:persona22}
\end{figure}

\begin{table}[t]
\centering
\scriptsize
\setlength{\tabcolsep}{3.5pt}
\begin{tabular*}{\linewidth}{@{\extracolsep{\fill}}l rrrrr rrrrr}
\toprule
& \multicolumn{5}{c}{As Seller} & \multicolumn{5}{c}{As Buyer} \\
\cmidrule(lr){2-6} \cmidrule(lr){7-11}
Model & Sales & Conv. & Ref. & Qty & Rating & Spend & Conv. & Ref. & Qty & Rating \\
\midrule
Qwen3-235B     & \$10,650 & 86.5 & 29.9 & 1.10 & 3.90 & \$11,325 & 87.3 & 22.7 & 1.11 & 3.68 \\
GPT-oss-120B   & \$10,283 & 74.3 & 18.2 & 1.09 & 3.71 &  \$8,323 & 68.8 & 19.9 & 1.12 & 3.73 \\
DeepSeek-V3.2  &  \$9,708 & 76.3 & 26.2 & 1.08 & 3.76 & \$11,425 & 81.3 & 23.6 & 1.11 & 3.62 \\
Gemini-3.1-Pro &  \$8,982 & 69.5 & 23.3 & 1.20 & 3.62 & \$10,167 & 84.5 & 29.0 & 1.17 & 3.89 \\
GPT-5.4        &  \$7,802 & 66.5 & 21.6 & 1.10 & 3.50 &  \$6,185 & 51.2 & 24.8 & 1.00 & 3.59 \\
\bottomrule
\end{tabular*}
\vspace{-0.25cm}
\caption{Aggregate behavioral outcomes of LLMs as sellers and buyers. Seller-side reports total sales, buyer-side reports total spending; both include conversion rate (Conv., \%), refund rate (Ref., \%), average quantity (Qty), and rating (1–5) from role-specific interactions.}
\vspace{-0.25cm}
\label{tab:llm_sales_spending}
\end{table}


\textbf{Results.} Figure \ref{fig:persona22} shows the inferred persona of five LLMs as seller and buyer roles. They exhibit a distinct combination of traits across roles. As sellers, most LLMs show relative assertive and friendly, except DeepSeek-V3.2 and GPT-5.4. Meanwhile, rationality also varies: the GPT family (GPT-oss-120B, GPT-5.4) is less emotional than others. As buyers, easygoing tendencies are generally prevalent across LLMs,  while differences emerge in price consciousness and rationality. Only Gemini-3.1-Pro exhibits price sensitivity, whereas other models are more price-indifferent. In terms of rationality, the GPT family exhibits more rational tendencies, consistent with the trends observed on the seller side. 

\vspace*{-0.1cm}
\subsection{Use Case 2: Revenue Impact of Seller--Buyer Interactions}
\label{sec:app_dynamics}
\vspace*{-0.1cm}

Building on the estimated LLM personas above, we examine how these inherent tendencies translate into economic outcomes when the six models interact as sellers and buyers. This analysis is critical, as retail outcomes are determined not only by individual behaviors but also by the interplay between seller and buyer personas.

\begin{wrapfigure}{r}{0.43\textwidth}
    \centering
    \vspace{-0.5cm}
    \includegraphics[width=\linewidth]{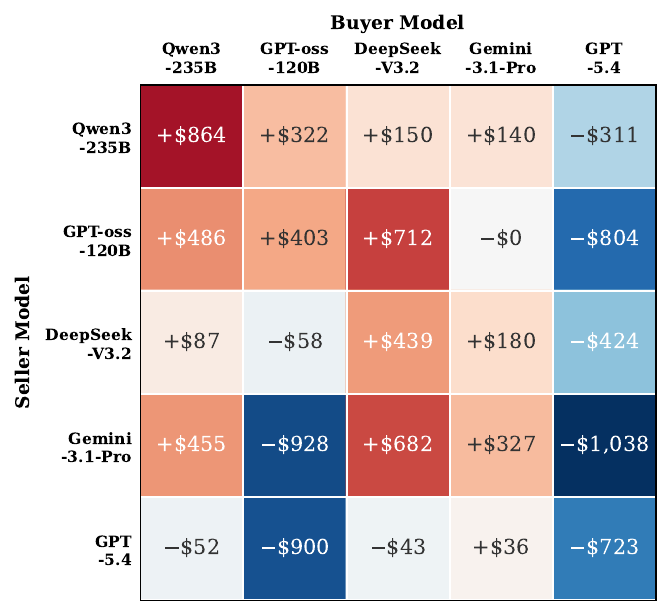}
    \vspace{-0.7cm}
    \caption{Normalized revenue heatmap across five LLMs.}
    \label{fig:seller_buyer_interaction}
    \vspace{-0.25cm}
\end{wrapfigure}

\textbf{Aggregate Behavioral Outcomes.} Table \ref{tab:llm_sales_spending} shows the aggregate behavioral outcomes of each LLM when acting as sellers and buyers in interactions with other LLMs, where seller-side metrics report the total sales across the 120 products each LLM sells, along with conversion rate (share of interactions resulting in purchase), refund rate (share of sold items that are refunded), average quantity per order, and average review rating, while buyer-side metrics report the total spending across the same set of interaction statistics computed from the buyer perspective. The results reveal clear and persona-consistent behavioral patterns across LLMs. As sellers, assertive and friendly models (\emph{e.g.}, Qwen3-235B, GPT-oss-120B) achieve higher sales and conversion, whereas more passive and reserved models (\emph{e.g.}, GPT-5.4) generate lower revenue and conversion. But it is of interest to see that emotionally-driven sellers (\emph{e.g.}, Qwen3-235B) show higher refund rates than more passive sellers (\emph{e.g.}, GPT-5.4). Meanwhile, friendly sellers appear to receive higher ratings, while easygoing buyers appear to give higher scores. 


\textbf{Pairwise Interaction Dynamics.} In contrast to the aggregate view, Figure \ref{fig:seller_buyer_interaction} explicitly captures pairwise interactions between LLMs, presenting a revenue heatmap where each cell shows the total net revenue when one LLM sells to another across the 120 products. For visual clarity, values are normalized by subtracting the mean revenue across all LLMs, where higher-than-average values are shown in red and lower-than-average values in blue. First, the top-selling LLM (Qwen3-235B) in Table \ref{tab:llm_sales_spending} does not consistently perform best across all buyers, indicating that seller--buyer persona pairing influences revenue outcomes. Second, interactions where the same LLM acts as both seller and buyer (self-pairing) do not consistently yield the highest revenue, indicating that pairing with the same model does not necessarily improve revenue outcomes. Third, Gemini-3.1-Pro shows highly polarized outcomes, performing well with easygoing buyers (\emph{e.g.}, Qwen3-235B, DeepSeek-V3.2) but poorly with rational ones (\emph{e.g.}, GPT-oss-120B, GPT-5.4). These findings underscore the value of our simulation framework, which enables controlled analysis of seller--buyer interactions and reveals interaction effects not captured by aggregate evaluations.

\vspace*{-0.15cm}
\subsection{Use Case 3: Evaluating Sales Strategy Effectiveness}
\label{sec:app_strategy}
\vspace*{-0.15cm}

Beyond user personas, revenue is also driven by the seller's product presentation strategy, particularly the sales script. As described in Section \ref{sec:method_framework}, our scripts are derived from real retail practices and reflect highly refined, human-designed strategies. We compare these with naive LLM-generated scripts (LLM-only, 0\%) and evaluate how revenue changes as human guidance is incrementally added to the prompt, introducing one key dimension at a time up to a fully guided setting (100\%). With partial guidance, a random subset of the four dimensions is provided per product. This highlights the ability of \algname{} to systematically assess the impact of high-quality sales strategies. To account for the impact of model capacity on script quality, we follow the same setup as in Section \ref{sec:application}, with the addition of Qwen3-80B/235B for script generation under varying prompts across five levels.

\begin{wraptable}{r}{0.45\textwidth}
\centering
\scriptsize
\setlength{\tabcolsep}{2.2pt}
\vspace{-0.3cm}
\begin{tabular}{l|c|cccc}
\toprule
 & & \multicolumn{4}{c}{Human Guidance Level} \\
\cmidrule(lr){3-6}
Seller Agent & LLM-Only & 25\% & 50\% & 75\% & 100\% \\
\midrule
Qwen3-80B  & \$6,246 & \$6,537 & \$6,830 & \$6,843 & \$6,707 \\
Qwen3-235B & \$6,330 & \$6,484 & \$6,376 & \$6,496 & \$6,695 \\
\midrule
Average   & \$6,288 & \$6,511 & \$6,603 & \$6,669 & \$6,701 \\
\bottomrule
\end{tabular}
\vspace{-0.2cm}
\caption{Revenue across different levels of human-guided sales strategies.}
\vspace{-0.25cm}
\label{tab:strategy_ablation}
\end{wraptable}
Table \ref{tab:strategy_ablation} presents the resulting revenue across different guidance levels and script-generation models, enabling a direct comparison between naive and progressively guided strategies. 
We first observe that larger models yield higher revenue under LLM-only settings; however, this gap becomes negligible with full human guidance ($100\%$) in script generation.
Revenue also increases in general as human guidance is progressively incorporated into the scripts. Importantly, the expected advantage of human-guided scripts is consistently reproduced in simulation with \algname{}, suggesting that our framework can serve as a controllable testbed for assessing which strategy components are associated with more consistent gains prior to deployment.


\vspace*{-0.15cm}
\section{Conclusion}
\label{sec:conclusion}
\vspace*{-0.2cm}

In this paper, we present \algname{}, a unified multi-stage simulation framework that captures how seller strategies, buyer personas, and interactions jointly drive retail outcomes, demonstrating both behavioral fidelity and economic consistency by reproducing key real-world patterns, while enabling practical, decision-oriented analysis such as persona inference and sales strategy evaluation, thereby establishing it as a controlled testbed for understanding and optimizing complex retail dynamics.

\paragraph{Ethics Statement.}
This work studies LLM-based retail simulation and raises no additional ethical concerns beyond standard human annotation practice. For human evaluation, we recruited annotators on Amazon Mechanical Turk with qualification filters (HIT approval rate over 90\%, at least 500 approved HITs, and a pre-task English comprehension threshold), and compensated annotators at \$7.50 per hour, above the U.S.\ federal minimum wage. We did not collect personally identifiable information at any stage. Experiments were conducted using commercial APIs and open-source models under their respective terms and licenses. Detailed model configuration details are provided in Appendix \ref{appendix:llm_configuration}, and annotation protocol and quality control are provided in Appendix \ref{appendix:human_annotation}.

\paragraph{Reproducibility Statement.}
We report the model backbones, checkpoints, and access routes used in all experiments, including proprietary models (Gemini-3-Flash, Gemini-3.1-Pro, GPT-5.4-mini, GPT-5.4) and open-source models (Qwen3-80B-A3B-Instruct, Qwen3-235B-A22B-Instruct, GPT-oss-120B, DeepSeek-V3.2). Proprietary models are accessed through their official APIs, while open-source models are deployed locally. To facilitate replication, we release human-annotation templates, detailed LLM configuration settings, prompt templates, and full simulation examples. Model configurations are provided in Appendix \ref{appendix:llm_configuration}; human-annotation templates and quality-control details are provided in Appendix \ref{appendix:human_annotation}; and prompts are provided in Appendix \ref{appendix:llm_prompt}.

\paragraph{Acknowledgments.}
This work was supported by the National Research Foundation of Korea (NRF) grant funded by the Korea government (MSIT) (No. RS-2024-00334343) and by Institute of Information \& communications Technology Planning \& Evaluation (IITP) grant funded by the Korea government (MSIT) (RS-2025-25464461, AI's Vision of Harmony: A Fair and Transparent Multimodal Agentic Platform for Conflict Mediation).

\bibliographystyle{assets/plainnat}
\bibliography{colm2026_conference}

@inproceedings{chu2025simul1,
  title     = {LLM-Based Multi-Agent System for Simulating and Analyzing Marketing and Consumer Behavior},
  author    = {Chu, Man-Lin and Terhorst, Lucian and Reed, Kadin and Ni, Tom and Chen, Weiwei and Lin, Rongyu},
  booktitle = {ICEBE},
  year      = {2025}
}

@inproceedings{zhang2025simulating,
  title     = {Simulating Classroom Education with LLM-Empowered Agents},
  author    = {Zhang, Zheyuan and Zhang-Li, Daniel and Yu, Jifan and Gong, Linlu and Zhou, Jinchang and Hao, Zhanxin and Jiang, Jianxiao and Cao, Jie and Liu, Huiqin and Liu, Zhiyuan and Hou, Lei and Li, Juanzi},
  booktitle = {NAACL},
  year      = {2025}
}

@inproceedings{li2024econagent,
  title     = {EconAgent: Large Language Model-Empowered Agents for Simulating Macroeconomic Activities},
  author    = {Li, Nian and Gao, Chen and Li, Mingyu and Li, Yong and Liao, Qingmin},
  booktitle = {ACL},
  year      = {2024}
}

@article{feng2025simcity,
  title   = {SimCity: Multi-Agent Urban Development Simulation with Rich Interactions},
  author  = {Feng, Yeqi and Lu, Yucheng and Su, Hongyu and Tao, Yixin and He, Tianxing},
  journal = {arXiv preprint arXiv:2510.01297},
  year    = {2025}
}

@inproceedings{yao2025taubench,
  title     = {$\tau$-bench: A Benchmark for Tool-Agent-User Interaction in Real-World Domains},
  author    = {Yao, Shunyu and Shinn, Noah and Razavi, Pedram and Narasimhan, Karthik},
  booktitle = {ICLR},
  year      = {2025}
}

@inproceedings{wang2025ecombench,
  title     = {ECom-Bench: Can LLM Agent Resolve Real-World E-commerce Customer Support Issues?},
  author    = {Wang, Haoxin and Peng, Xianhan and Cheng, Huang and Huang, Yizhe and Gong, Ming and Yang, Chenghan and Liu, Yang and Lin, Jiang},
  booktitle = {EMNLP},
  year      = {2025}
}

@inproceedings{zhou2026mixecom,
  title     = {Mix-Ecom: Towards Mixed-Type E-Commerce Dialogues with Complex Domain Rules},
  author    = {Zhou, Chenyu and Shi, Xiaoming and Qiu, Hui and Jiang, Yankai and Liu, ShaoGuo and Gao, Tingting and Leng, Haitao and Zheng, Xiawu and Ji, Rongrong},
  booktitle = {ICLR},
  year      = {2026}
}

@inproceedings{hu2026simbench,
  title     = {SimBench: Benchmarking the Ability of Large Language Models to Simulate Human Behaviors},
  author    = {Hu, Tiancheng and Baumann, Joachim and Lupo, Lorenzo and Collier, Nigel and Hovy, Dirk and R{\"o}ttger, Paul},
  booktitle = {ICLR},
  year      = {2026}
}

@inproceedings{zhao2026ecomstage,
  title     = {EComStage: Stage-wise and Orientation-specific Benchmarking for Large Language Models in E-commerce},
  author    = {Zhao, Kaiyan and Meng, Zijie and Xie, Zheyong and Duan, Jin and Hu, Yao and Liu, Zuozhu and Cao, Shaosheng},
  booktitle = {ICLR},
  year      = {2026}
}

@article{wang2025customerr1,
  title   = {Customer-R1: Personalized Simulation of Human Behaviors via RL-based LLM Agent in Online Shopping},
  author  = {Wang, Ziyi and Lu, Yuxuan and Zhang, Yimeng and Huang, Jing and Wang, Dakuo},
  journal = {arXiv preprint arXiv:2510.07230},
  year    = {2025}
}

@inproceedings{zhang2026shopr1,
  title     = {SHOP-R1: Rewarding LLMs to Simulate Human Behavior in Online Shopping via Reinforcement Learning},
  author    = {Zhang, Yimeng and Wang, Tian and Gesi, Jiri and Wang, Ziyi and Lu, Yuxuan and Lin, Jiacheng and Zhan, Sinong and Gao, Vianne and Jiao, Ruochen and Liu, Junze and Qian, Kun and Tang, Yuxin and Xue, Ran and Zhang, Houyu and Cui, Qingjun and Guo, Yufan and Wang, Dakuo},
  booktitle = {ICLR},
  year      = {2026}
}

@article{bansal2025magenticmarketplace,
  title   = {Magentic Marketplace: An Open-Source Environment for Studying Agentic Markets},
  author  = {Bansal, Gagan and Hua, Wenyue and Huang, Zezhou and Fourney, Adam and Swearngin, Amanda and Epperson, Will and Payne, Tyler and Hofman, Jake M. and Lucier, Brendan and Singh, Chinmay and Mobius, Markus and Nambi, Akshay and Yadav, Archana and Gao, Kevin and Rothschild, David M. and Slivkins, Aleksandrs and Goldstein, Daniel G. and Mozannar, Hussein and Immorlica, Nicole and Murad, Maya and Vogel, Matthew and Kambhampati, Subbarao and Horvitz, Eric and Amershi, Saleema},
  journal = {arXiv preprint arXiv:2510.25779},
  year    = {2025}
}

@article{wang2025opera,
  title   = {OPeRA: A Dataset of Observation, Persona, Rationale, and Action for Evaluating LLMs on Human Online Shopping Behavior Simulation},
  author  = {Wang, Ziyi and Lu, Yuxuan and Li, Wenbo and Amini, Amirali and Sun, Bo and Bart, Yakov and Lyu, Weimin and Gesi, Jiri and Wang, Tian and Huang, Jing and Su, Yu and Ehsan, Upol and Alikhani, Malihe and Li, Toby Jia-Jun and Chilton, Lydia and Wang, Dakuo},
  journal = {arXiv preprint arXiv:2506.05606},
  year    = {2025}
}

@article{mansour2025paars,
  title   = {PAARS: Persona Aligned Agentic Retail Shoppers},
  author  = {Mansour, Saab and Perelli, Leonardo and Mainetti, Lorenzo and Davidson, George and D'Amato, Stefano},
  journal = {arXiv preprint arXiv:2503.24228},
  year    = {2025}
}

@inproceedings{wang2025ecomscriptbench,
  title     = {EcomScriptBench: A Multi-task Benchmark for E-commerce Script Planning via Step-wise Intention-Driven Product Association},
  author    = {Wang, Weiqi and Cui, Limeng and Liu, Xin and Nag, Sreyashi and Xu, Wenju and Luo, Chen and Sarwar, Sheikh Muhammad and Li, Yang and Gu, Hansu and Liu, Hui and Yu, Changlong and Bai, Jiaxin and Gao, Yifan and Zhang, Haiyang and He, Qi and Ji, Shuiwang and Song, Yangqiu},
  booktitle = {ACL},
  year      = {2025}
}

@inproceedings{bianchi2024negotiationarena,
  title     = {How Well Can {LLMs} Negotiate? {NEGOTIATIONARENA} Platform and Analysis},
  author    = {Bianchi, Federico and Chia, Patrick John and Yuksekgonul, Mert and
               Tagliabue, Jacopo and Jurafsky, Dan and Zou, James},
  booktitle = {ICML},
  year      = {2024}
}

@inproceedings{xia2024bargaining,
  title     = {Measuring Bargaining Abilities of {LLMs}: A Benchmark and A Buyer-Enhancement Method},
  author    = {Xia, Tian and He, Zhiwei and Ren, Tong and Miao, Yibo and
               Zhang, Zhuosheng and Yang, Yang and Wang, Rui},
  booktitle = {ACL},
  year      = {2024}
}

@article{wang2025shoppingbench,
  title   = {{ShoppingBench}: A Real-World Intent-Grounded Shopping Benchmark
             for {LLM}-based Agents},
  author  = {Wang, Jiangyuan and Xiao, Kejun and Sun, Qi and Zhao, Huaipeng and
             Luo, Tao and Zhang, Jian Dong and Zeng, Xiaoyi},
  journal = {arXiv preprint arXiv:2508.04266},
  year    = {2025}
}

@inproceedings{hwang2026personality,
  title     = {Personality Editing for Language Models through Adjusting Self-Referential Queries},
  author    = {Hwang, Seojin and Kim, Yumin and Kim, Byeongjeong and Shin, Donghoon and Lee, Hwanhee},
  booktitle = {EACL},
  year      = {2026}
}

@inproceedings{majumder2020hiking,
  title     = {Like Hiking? You Probably Enjoy Nature: Persona-grounded Dialog with Commonsense Expansions},
  author    = {Majumder, Bodhisattwa Prasad and Jhamtani, Harsh and Berg-Kirkpatrick, Taylor and McAuley, Julian},
  booktitle = {EMNLP},
  year      = {2020}
}

@article{ahearne2022future,
  title={The future of buyer--seller interactions: A conceptual framework and research agenda},
  author={Ahearne, Michael and Atefi, Yashar and Lam, Son K and Pourmasoudi, Mohsen},
  journal={Journal of the Academy of Marketing Science},
  volume={50},
  number={1},
  pages={22--45},
  year={2022}
}

@article{mukherjee2021webrooming,
  title={Webrooming and showrooming: a multi-stage consumer decision process},
  author={Mukherjee, Srabanti and Chatterjee, Swagato},
  journal={Marketing Intelligence \& Planning},
  year={2021}
}

@inproceedings{shi2025you,
  title={You are what you bought: Generating customer personas for e-commerce applications},
  author={Shi, Yimin and Fei, Yang and Zhang, Shiqi and Wang, Haixun and Xiao, Xiaokui},
  booktitle={SIGIR},
  year={2025}
}

@article{li2025consumer,
  title={Consumer attention and market concentration in e-commerce: an agent-based perspective},
  author={Li, Teng and Wang, Shaoni and Zhou, Deyi and Razzaq, Amar},
  journal={Journal of Economic Interaction and Coordination},
  volume={20},
  number={4},
  pages={959--985},
  year={2025}
}

@article{gupta2026customer,
  title={Customer journey and experience in multi-stage e-retail delivery system (e-RDS): a process--perception perspective},
  author={Gupta, Piyush and Singh, Jagroop and Sachan, Amit and Srivastava, Abhishek},
  journal={Journal of Global Operations and Strategic Sourcing},
  pages={1--21},
  year={2026}
}

@article{lemon2016understanding,
  title={Understanding customer experience throughout the customer journey},
  author={Lemon, Katherine N and Verhoef, Peter C},
  journal={Journal of Marketing},
  volume={80},
  number={6},
  pages={69--96},
  year={2016}
}

@incollection{phillips2021pricing,
  title={Pricing and revenue optimization},
  author={Phillips, Robert L},
  booktitle={Pricing and Revenue Optimization},
  year={2021}
}

@inproceedings{wang2025ecom,
  title={ECom-Bench: Can LLM Agent Resolve Real-World E-commerce Customer Support Issues?},
  author={Wang, Haoxin and Peng, Xianhan and Cheng, Huang and Huang, Yizhe and Gong, Ming and Yang, Chenghan and Liu, Yang and Lin, Jiang},
  booktitle={EMNLP},
  pages={276--284},
  year={2025}
}

@inproceedings{ban2026completing,
  title={Completing Missing Annotation: Multi-Agent Debate for Accurate and Scalable Relevant Assessment for IR Benchmarks},
  author={Ban, Minjeong and Choi, Jeonghwan and Min, Hyangsuk and Kim, Nicole Hee-Yeon and Kim, Minseok and Lee, Jae-Gil and Song, Hwanjun},
  booktitle={ICLR},
  year={2026}
}

@inproceedings{cook2025efficient,
  title={Efficient Annotator Reliability Assessment with EffiARA},
  author={Cook, Owen and Vasilakes, Jake A and Roberts, Ian and Song, Xingyi},
  booktitle={ACL},
  year={2025}
}

@article{gao2024large,
  title={Large language models empowered agent-based modeling and simulation: A survey and perspectives},
  author={Gao, Chen and Lan, Xiaochong and Li, Nian and Yuan, Yuan and Ding, Jingtao and Zhou, Zhilun and Xu, Fengli and Li, Yong},
  journal={Humanities and Social Sciences Communications},
  volume={11},
  number={1},
  pages={1--24},
  year={2024}
}

@article{hou2024bridging,
  title={Bridging Language and Items for Retrieval and Recommendation},
  author={Hou, Yupeng and Li, Jiacheng and He, Zhankui and Yan, An and Chen, Xiusi and McAuley, Julian},
  journal={arXiv preprint arXiv:2403.03952},
  year={2024}
}

@article{ding2023impact,
  title={The impact of product diversity and distribution networks on consumption expansion},
  author={Ding, Chaoxun and Zhang, Ruidan and Wu, Xuepin},
  journal={Journal of Business Research},
  volume={161},
  pages={113833},
  year={2023}
}

@article{edelman2015competing,
  title={Competing on customer journeys},
  author={Edelman, David C and Singer, Marc},
  journal={Harvard Business Review},
  volume={93},
  number={11},
  pages={88--100},
  year={2015}
}

@article{almansoori2025self,
  title={Self-evolving multi-agent simulations for realistic clinical interactions},
  author={Almansoori, Mohammad and Kumar, Komal and Cholakkal, Hisham},
  journal={arXiv preprint arXiv:2503.22678},
  year={2025}
}

@article{sturley2018evaluating,
  title={Evaluating the potential of agent-based modelling to capture consumer grocery retail store choice behaviours},
  author={Sturley, Charlotte and Newing, Andy and Heppenstall, Alison},
  journal={The International Review of Retail, Distribution and Consumer Research},
  volume={28},
  number={1},
  pages={27--46},
  year={2018}
}

@article{jamali2026agent,
  title={Agent-based modeling and simulation for economic markets: a comprehensive review of applications, challenges, and opportunities},
  author={Jamali, Ruhollah and Lazarova-Molnar, Sanja},
  journal={Journal of Simulation},
  pages={1--50},
  year={2026}
}

@article{kanwal2021gender,
  title   = {Systematic Review of Gender Differences and Similarities in Online Consumers' Shopping Behavior},
  author  = {Kanwal, Maria and Burki, Umar and Ali, Raza and Dahlstrom, Robert},
  journal = {Journal of Consumer Marketing},
  year    = {2021}
}

@article{redine2022impulse,
  title   = {Impulse Buying: A Systematic Literature Review and Future Research Directions},
  author  = {Redine, Artem and Deshpande, Sameer and Jebarajakirthy, Charles and Surachartkumtonkun, Jiraporn},
  journal = {International Journal of Consumer Studies},
  year    = {2022}
}

@article{cakici2022mediating,
  title   = {The Mediating Effect of Consumers' Price Level Perception and Emotions towards Supermarkets},
  author  = {Cakici, A. Celil and Tekeli, Sena},
  journal = {European Journal of Management and Business Economics},
  year    = {2022}
}

@article{iyer2020impulse,
  title   = {Impulse Buying: A Meta-Analytic Review},
  author  = {Iyer, Gopalkrishnan R. and Blut, Markus and Xiao, Sarah Hong and Grewal, Dhruv},
  journal = {Journal of the Academy of Marketing Science},
  year    = {2020}
}

@article{chaker2024quiet,
  title   = {Quiet Sellers: When Introversion Drives Salesperson Performance},
  author  = {Chaker, Nawar N. and Habel, Johannes and Hartmann, Nathaniel N. and Johannsen, Felix and Rusche, Heinrich},
  journal = {Journal of Retailing},
  year    = {2024}
}

@article{wongkitrungrueng2020livestreaming,
  title   = {The Role of Live Streaming in Building Consumer Trust and Engagement with Social Commerce Sellers},
  author  = {Wongkitrungrueng, Apiradee and Assarut, Nuttapol},
  journal = {Journal of Business Research},
  year    = {2020}
}

@article{cheng2021picky,
  title   = {Identifying Picky Shoppers: Who They Are and How to Spot Them},
  author  = {Cheng, Andong and Baumgartner, Hans and Meloy, Margaret G.},
  journal = {Journal of Consumer Psychology},
  year    = {2021}
}

@article{nowlin2018emotion,
  title   = {Emotion in Sales Performance: Affective Orientation and Need for Cognition and the Mediating Role of Motivation to Work},
  author  = {Nowlin, Edward and Walker, Danielle and Deeter-Schmelz, Dawn R. and Haas, Alexander},
  journal = {Journal of Business \& Industrial Marketing},
  year    = {2018}
}

@article{hitsch2021prices,
  title   = {Prices and Promotions in U.S. Retail Markets},
  author  = {Hitsch, G{\"u}nter J. and Horta\c{c}su, Ali and Lin, Xiliang},
  journal = {Quantitative Marketing and Economics},
  year    = {2021}
}

@article{openai2026gpt54,
  title={GPT-5.4 Thinking System Card},
  author={{OpenAI}},
  year={2026},
  month={March},
}

@article{google2026geminipro,
  title={Gemini 3.1 Pro Model Card},
  author={{Google DeepMind}},
  year={2026},
  month={Feb},
}

@article{google2025geminiflash,
  title={Gemini 3 Flash Model Card},
  author={{Google DeepMind}},
  year={2025},
  month={December},
}

@article{openai2025gpt5,
  title={GPT-5 System Card},
  author={{OpenAI}},
  year={2025},
  month={August},
}

@article{liu2025deekseek,
  title   = {DeepSeek-V3.2: Pushing the Frontier of Open Large Language Models},
  author  = {DeepSeek-AI and Liu, Aixin and Mei, Aoxue and others},
  journal = {arXiv preprint arXiv:2512.02556},
  year    = {2025}
}

@article{yang2025qwen3,
  title={Qwen3 Technical Report},
  author={Yang, An and others},
  journal={arXiv preprint arXiv:2505.09388},
  year={2025}
}

@article{zhu2025automatedrisky,
  title     = {The Automated but Risky Game: Modeling Agent-to-Agent Negotiations and Transactions in Consumer Markets},
  author    = {Zhu, Shenzhe and Sun, Jiao and Nian, Yi and South, Tobin and
               Pentland, Alex and Pei, Jiaxin},
  journal={arXiv preprint arXiv:2506.00073},
  year      = {2025}
}

\appendix
\clearpage

\begin{table}[!ht]
\scriptsize
\centering
\vspace{-0.4cm}
\setlength{\tabcolsep}{4pt}
\begin{tabular}{lllll}
\toprule
Type & Model & Model Checkpoint & Source & Reference \\
\midrule
\multirow{4}{*}{Open-source} 
& Qwen3-80B-A3B-Inst. & \texttt{Qwen/Qwen3-Next-80B-A3B-Instruct} & HuggingFace & \citet{yang2025qwen3} \\
& Qwen3-235B-A22B-Inst. & \texttt{Qwen/Qwen3-235B-A22B-Instruct-2507} & HuggingFace & \citet{yang2025qwen3} \\
& GPT-oss-120B & \texttt{openai/gpt-oss-120b} & HuggingFace & \citet{openai2025gpt5} \\
& DeepSeek-V3.2 & \texttt{deepseek-ai/DeepSeek-V3.2} & HuggingFace & \citet{liu2025deekseek} \\
\midrule
\multirow{4}{*}{Proprietary} 
& Gemini-3-Flash & \texttt{gemini-3-flash-preview} & Google API & \citet{google2025geminiflash} \\
& Gemini-3.1-Pro & \texttt{gemini-3.1-pro-preview} & Google API & \citet{google2026geminipro} \\
& GPT-5.4-mini & \texttt{gpt-5.4-mini-2026-03-17} & OpenAI API & \citet{openai2026gpt54} \\
& GPT-5.4 & \texttt{gpt-5.4-2026-03-05} & OpenAI API & \citet{openai2026gpt54} \\
\bottomrule
\end{tabular}
\vspace{-0.25cm}
\caption{Backbone LLM configurations for data processing and simulating with \algname{}.}
\vspace{-0.3cm}
\label{tab:backbone_llm}
\end{table}

\begin{table*}[!ht]
\scriptsize
\centering
\begin{tabularx}{\textwidth}{l l 
    >{\hsize=1.1\hsize\centering\arraybackslash}X 
    >{\hsize=1.1\hsize\centering\arraybackslash}X 
    >{\hsize=0.95\hsize\centering\arraybackslash}X 
    >{\hsize=0.95\hsize\centering\arraybackslash}X 
    >{\hsize=0.9\hsize\centering\arraybackslash}X 
    >{\hsize=1.0\hsize\centering\arraybackslash}X 
    >{\hsize=1.0\hsize\centering\arraybackslash}X 
    >{\hsize=0.9\hsize\centering\arraybackslash}X} 
\toprule
\multirow{3}{*}{Source} & \multirow{3}{*}{Model} & \multicolumn{2}{c}{Pricing} & \multicolumn{3}{c}{Non-purchase} & \multicolumn{3}{c}{Purchase} \\
\cmidrule(lr){5-7} \cmidrule(lr){8-10}
& & \multicolumn{2}{c}{(\$ / 1M tokens)} & \multicolumn{2}{c}{Tokens} & \multirow{2}{*}{\shortstack{Cost}} & \multicolumn{2}{c}{Tokens} & \multirow{2}{*}{\shortstack{Cost}} \\
\cmidrule(lr){5-6} \cmidrule(lr){8-9}
& & In & Out & In & Out & & In & Out & \\
\midrule
\multirow{4}{*}{OpenRouter}
& Qwen3-80B-A3B-Inst. & \$ 0.090 & \$ 1.10 & 17,515 & 2,092 & 0.0039 & 27,098 & 2,773 & 0.0055 \\
& Qwen3-235B-A22B-Inst. & \$ 0.071 & \$ 0.10 & 20,836 & 2,509 & 0.0017 & 32,609 & 3,376 & 0.0027 \\
& GPT-oss-120B & \$ 0.039 & \$ 0.19 & 38,462 & 3,580 & 0.0022 & 62,930 & 5,346 & 0.0035 \\
& DeepSeek-V3.2 & \$ 0.260 & \$ 0.38 & 28,331 & 2,456 & 0.0083 & 43,989 & 3,338 & 0.0127 \\
\midrule
\multirow{2}{*}{Google API}
& Gemini-3-Flash & \$ 0.500 & \$ 3.00 & 27,720 & 2,787 & 0.0222 & 39,623 & 3,581 & 0.0306 \\
& Gemini-3.1-Pro & \$ 2.000 & \$ 12.00 & 23,766 & 2,590 & 0.0786 & 36,151 & 3,484 & 0.1141 \\
\midrule
\multirow{2}{*}{OpenAI API}
& GPT-5.4-mini & \$ 0.750 & \$ 4.50 & 25,752 & 2,467 & 0.0304 & 42,636 & 3,011 & 0.0455 \\
& GPT-5.4 & \$ 2.500 & \$ 15.00 & 36,207 & 3,367 & 0.1410 & 54,567 & 4,383 & 0.2022 \\
\bottomrule
\end{tabularx}
\vspace{-0.25cm}
\caption{Token consumption and API cost per simulation across eight LLMs.}
\vspace{-0.3cm}
\label{tab:final_cost_updated_exp4}
\end{table*}

\section{LLM Configurations and Simulation Cost}
\label{appendix:llm_configuration}

\paragraph{LLM Configuration.}
Table \ref{tab:backbone_llm} summarizes the LLMs used as backbone models for product data preprocessing and simulating seller and buyer agents in \algname{}. For proprietary models including Gemini-3-Flash \citep{google2025geminiflash} Gemini-3.1-Pro \citep{google2026geminipro}, GPT-5.4-mini and GPT-5.4 \citep{openai2026gpt54}, we access them via their respective APIs. For open-source models including Qwen3-80B-A3B-Instruct and Qwen3-235B-A22B-Instruct \citep{yang2025qwen3}, GPT-oss-120B \citep{openai2025gpt5} and DeepSeek-V3.2 \citep{liu2025deekseek}, we deploy them locally on NVIDIA RTX PRO 6000 Blackwell Server Edition GPUs (96GB VRAM). All experiments are conducted with the default temperature setting of each model, 
and all models receive a unified prompt format within each experimental condition 
to facilitate fair cross-model comparisons.

\paragraph{Simulation Cost.}
Table \ref{tab:final_cost_updated_exp4} reports token consumption and estimated API cost averaged over 96 simulations per model. For commercial models, costs are based on each provider's pricing; for open-source models, costs are estimated using OpenRouter as a proxy for API-based access, and do not reflect the GPU-based setting used in our experiments. Non-purchase cases are shorter than purchase cases due to the absence of the post-purchase CS and review stage. Within each simulation, input token counts vary across models as a function of conversation length; since the full conversation history is passed at each turn, models that generate longer responses or sustain more dialogue turns accumulate larger inputs in subsequent turns, leading to model-level variance in token consumption.

\section{Sales Strategy}
\label{appendix:sales_strategy}

\paragraph{Two-step Design Rationale.}
The persuasion pipeline separates strategy formulation from script generation. In the first step, the seller agent derives a structured sales strategy conditioned on product information and its assigned persona. In the second step, the strategy serves as a blueprint for generating the persuasion script, ensuring that key messaging priorities are explicitly planned before being translated into script.

\paragraph{Industry Grounding.}
To ensure simulation fidelity, the strategy framework is grounded in authentic professional practice. We analyzed seven high-quality sales scripts provided by a professional retail company and identified four recurring strategic patterns and a consistent five-part sectional flow: opening hook, three core persuasion points covering target expansion, value validation, and objection handling, and a closing call-to-action. These observations directly informed the four strategy dimensions and script structure used in \algname{}, making seller behavior more likely to reflect commercially viable persuasion patterns. The four strategy dimensions are defined as follows.

\begin{enumerate}[leftmargin=*, noitemsep, topsep=0pt]
\item \textbf{Target Expansion.} The seller defines a core audience based on demographics, lifestyle, and needs, then identifies secondary audiences (\emph{e.g.}, gift buyers or multi-purpose users) to broaden reach.
\item \textbf{Value Proposition.} The seller identifies key buying factors, translates product attributes into practical or emotional benefits, and selects a persuasion method, namely \emph{Competitive Comparison}, \emph{Technical Authority}, or \emph{Value-per-Use} logic, to justify price--quality trade-offs.
\item \textbf{Contextual Urgency.} The seller establishes a concrete reason to buy now, grounded in seasonal timing, usage context, or product-specific milestones.
\item \textbf{Objection Handling.} The seller anticipates likely buyer hesitations and prepares persuasive responses that turn concerns into purchase motivation.
\end{enumerate}

\paragraph{Strategy-to-Script Transition.}
The generated strategy is passed as input to the script generation stage, where the seller constructs a five-part persuasion script. Throughout both stages, the seller's persona traits, namely assertiveness, friendliness, and rationality, govern strategic emphasis, ensuring that different seller agents yield distinct yet coherent sales approaches for the same product. In non-persona settings, explicit trait constraints are removed from the prompt as described in Appendix \ref{appendix:llm_prompt}, and the model defaults to its inherent style while maintaining the same stage structure and task requirements. A concrete simulation example is provided in Appendix \ref{appendix:simulation_example}.

\section{Human Annotation}
\label{appendix:human_annotation}

\paragraph{Annotator Qualification and Compensation.}
Since RetailSim aims to reproduce real-world retail dynamics and support prediction of downstream outcomes, the quality of human judgments plays an important role in validating simulation fidelity. We thus collect human annotations via Amazon Mechanical Turk (MTurk), restricting participation to annotators who meet the following qualifications: a HIT approval rate above 90\%, at least 500 approved HITs, and a minimum score of 90 on a custom English comprehension exam administered prior to the task. Annotators are compensated at \$7.50 per hour, exceeding the U.S. federal minimum wage. No personally identifiable information is collected throughout the annotation process.

\begin{figure}[!p]
    \centering
    \includegraphics[width=1\textwidth]{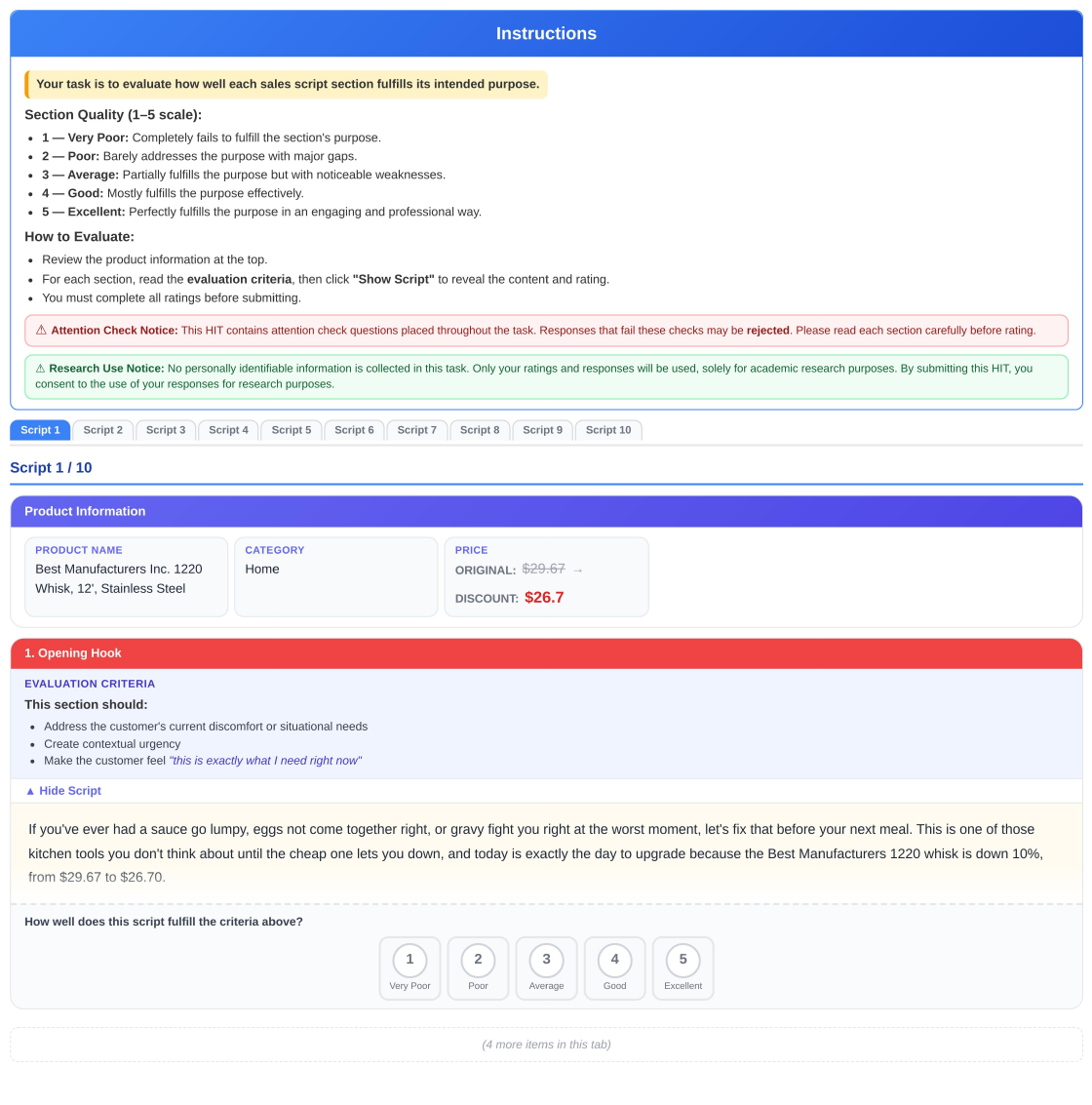}
    \vspace{-1cm}
    \caption{Example of annotation template for sales script quality evaluation (1–5 Likert scale). Each HIT contains 10 scripts with 5 sections each.}
    \vspace{-0.3cm}
    \label{fig:broadcast_template}
\end{figure}

\begin{figure}[!p]
    \centering
    \includegraphics[width=1\textwidth]{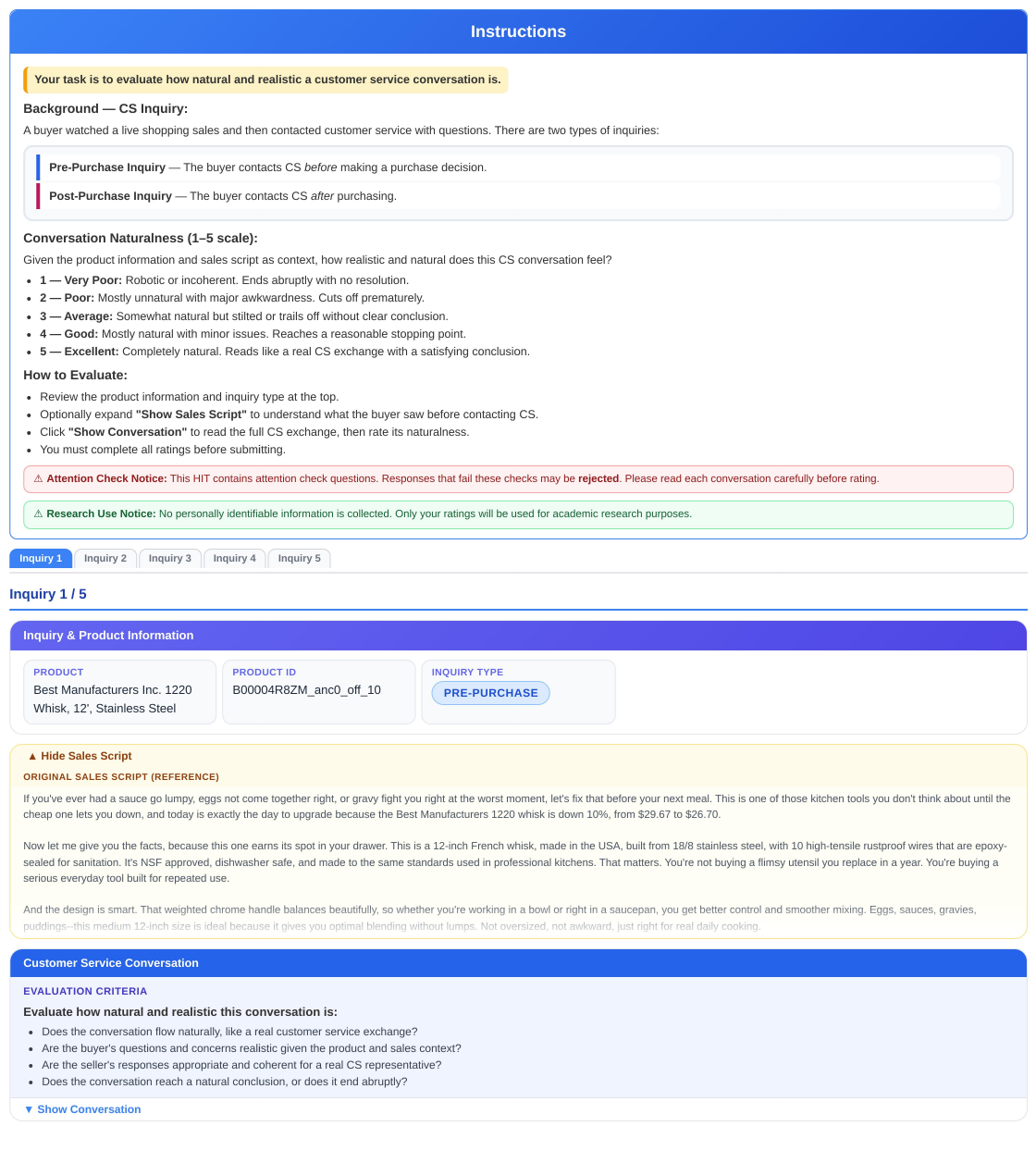}
    \vspace{-1cm}
    \caption{Example of annotation template for pre- and post-purchase inquiry naturalness evaluation (1–5 Likert scale). Each HIT contains 5 inquiries.}
    \vspace{-0.3cm}
    \label{fig:csinquiry_template}
\end{figure}

\begin{figure}[!p]
    \centering
    \includegraphics[width=1\textwidth]{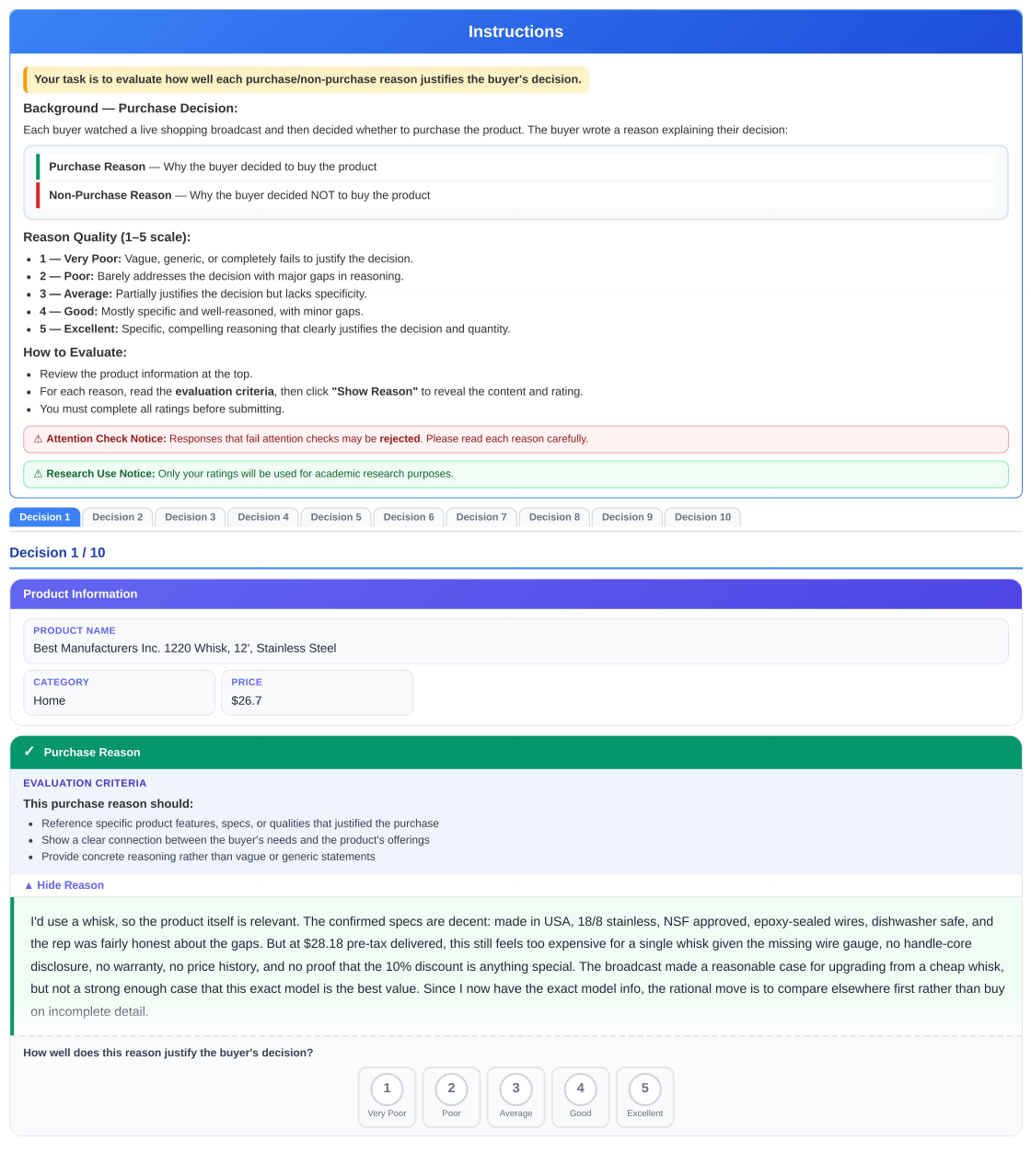}
    \vspace{-1cm}
    \caption{Example of annotation template for purchase/non-purchase reason quality evaluation (1–5 Likert scale). Each HIT contains 10 decisions.}
    \vspace{-0.3cm}
    \label{fig:purchase_template}
\end{figure}

\begin{figure}[!p]
    \centering
    \includegraphics[width=1\textwidth]{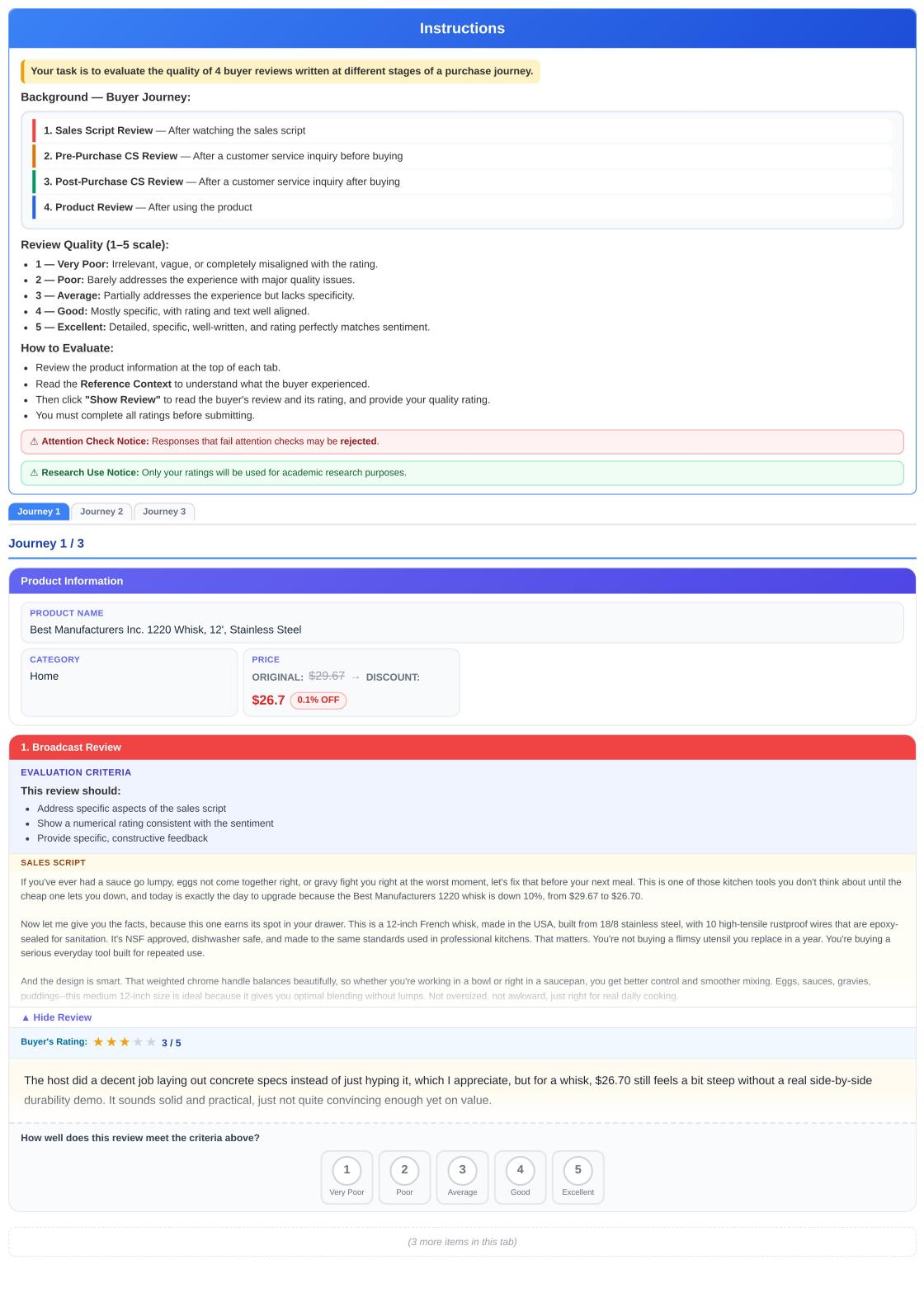}
    \vspace{-1cm}
    \caption{Example of annotation template for buyer review quality evaluation across four journey stages (1–5 Likert scale). Each HIT contains 3 journeys.}
    \vspace{-0.3cm}
    \label{fig:review_template}
\end{figure}

\begin{figure}[!p]
    \centering
    \includegraphics[width=1\textwidth]{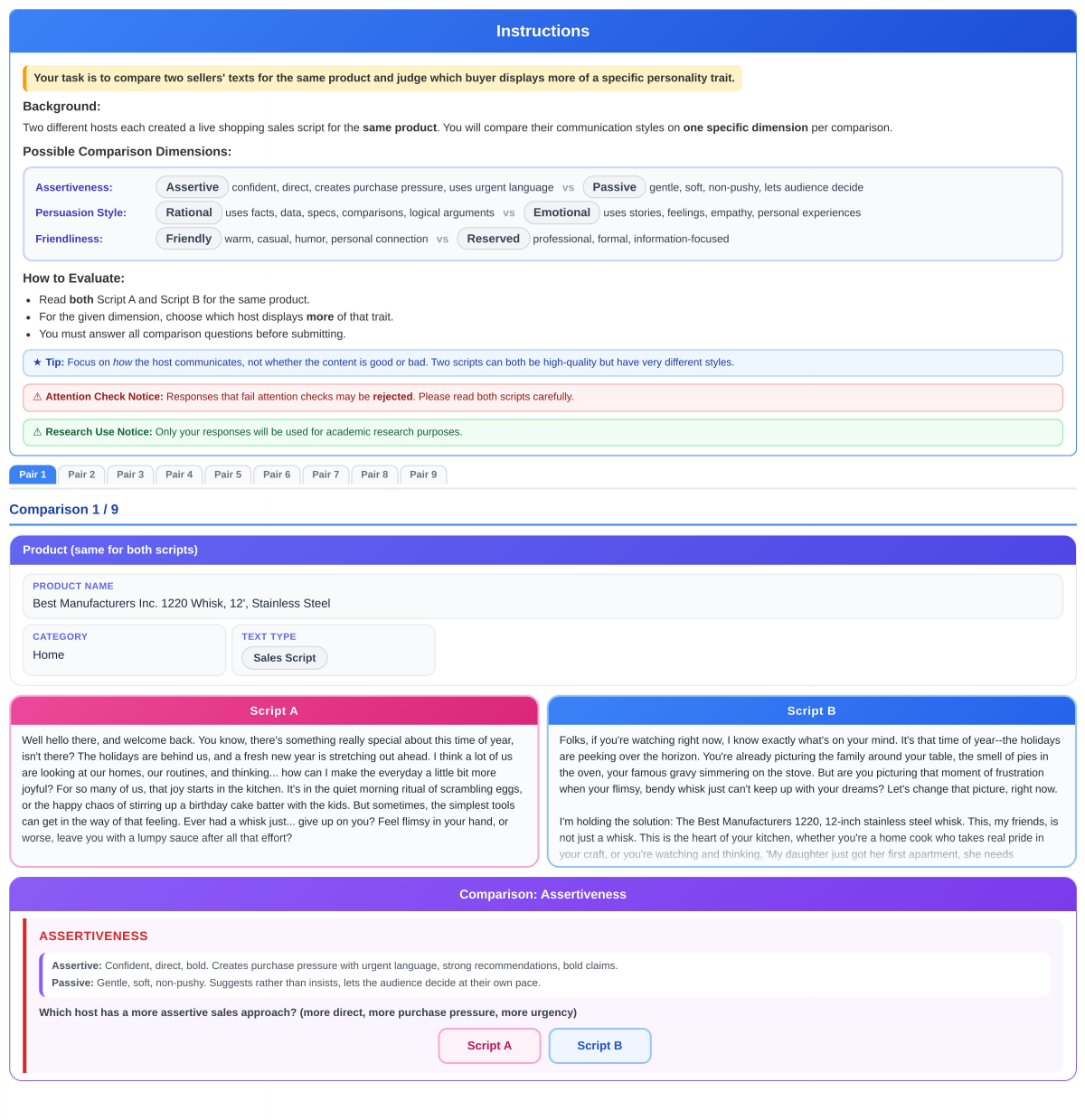}
    \vspace{-1cm}
    \caption{Example of annotation template for pairwise seller persona comparison (\emph{assertiveness}, \emph{friendliness}, and \emph{rationality}). Each HIT contains 9 pairs.}
    \vspace{-0.3cm}
    \label{fig:seller_template}
\end{figure}

\begin{figure}[!p]
    \centering
    \includegraphics[width=1\textwidth]{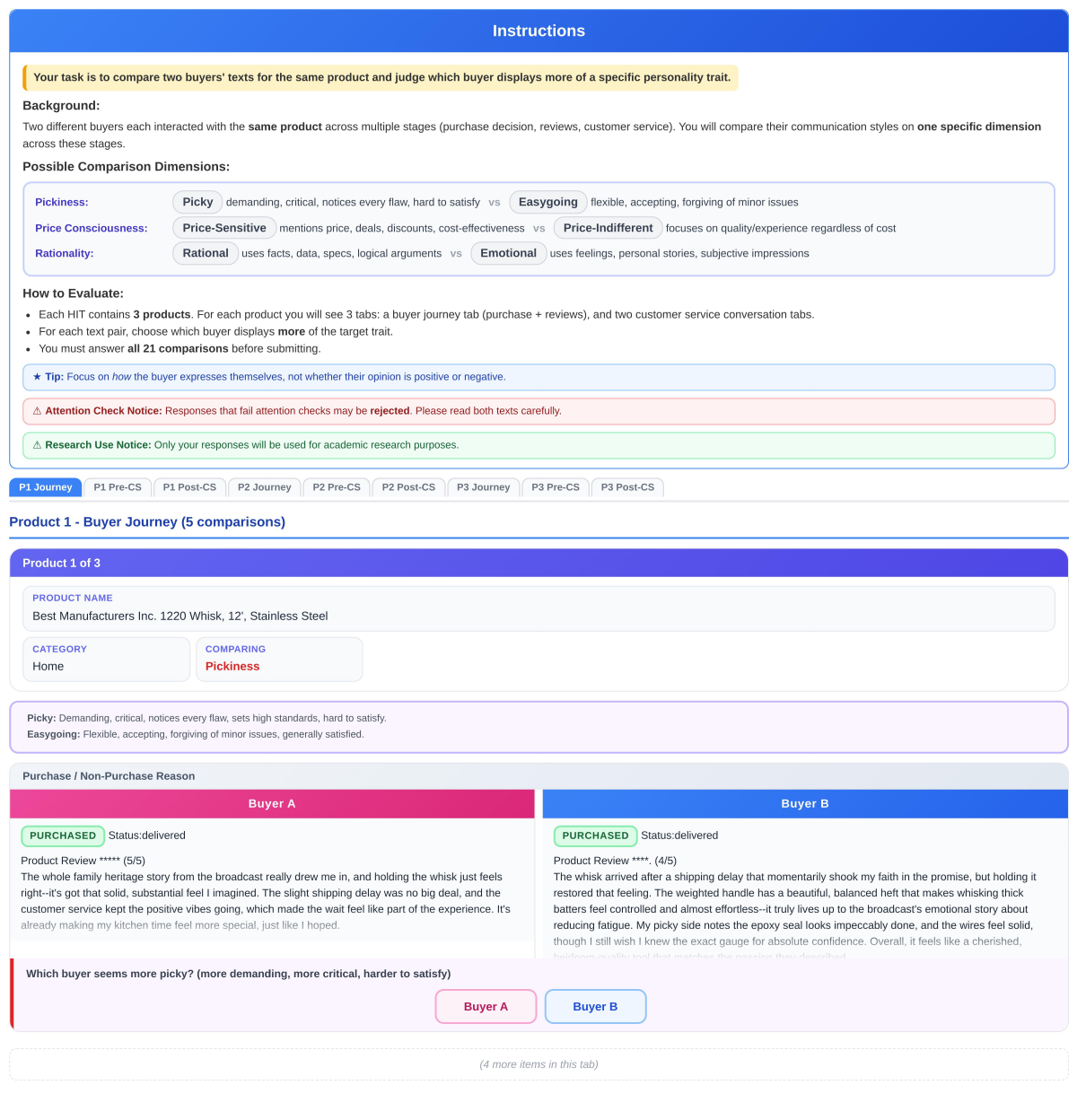}
    \vspace{-1cm}
    \caption{Example of annotation template for pairwise buyer persona comparison (\emph{pickiness}, \emph{price consciousness}, \emph{rationality}). Each HIT contains 3 products with 21 comparisons.}
    \vspace{-0.3cm}
    \label{fig:buyer_template}
\end{figure}

\paragraph{Task Design.}
We design two complementary annotation tasks targeting stage-level evaluation of simulation fidelity.
\begin{itemize}[leftmargin=*, noitemsep, topsep=0pt]
\item \emph{Task Fidelity}: To assess how faithfully each stage is simulated, we present annotators with stage-specific evaluation criteria reflecting the intended behaviors and objectives of each stage, and ask them to rate simulation outputs accordingly. Concretely, annotators evaluate outputs across five pipeline stages (persuasive scripts, pre-purchase inquiries, purchase decisions, post-purchase inquiries, and product reviews) on a 5-point Likert scale (1 being Very Poor and 5 being Excellent). We collect human annotation on 12 products across all 8 LLM backbones, yielding $8 \times 12 \times 5 \times 3 = 1{,}440$ simulations in total. As each stage involves distinct simulation instructions and evaluation criteria, we design a separate MTurk template per stage, as shown in Figures \ref{fig:broadcast_template}--\ref{fig:review_template}.

\item \emph{Persona Fidelity}: Annotators are presented with A/B pairs of simulations generated under opposing trait conditions (e.g., rational vs.\ emotional) and asked to identify which simulation output better reflects the target trait, while all other traits are held fixed. For each of the 6 persona traits across 2 agent roles, we construct 4 A/B pairs per product per LLM backbone (one for each combination of the remaining two traits), yielding $8 \times 12 \times 2 \times 3 \times 4 = 2{,}304$ pairs in total. As seller and buyer agents involve distinct trait definitions and output types, we design separate MTurk templates for each role, as shown in Figures \ref{fig:seller_template}--\ref{fig:buyer_template}.
\end{itemize}

\paragraph{Quality Control.}
To maintain annotation reliability, each simulation output and A/B pair is evaluated by three independent annotators. For task fidelity, we report the average score across the three annotators, while for persona fidelity, the final label is determined by majority vote. We measure inter-annotator agreement (IAA) using Krippendorff's $\alpha$, where $\alpha \geq 0.67$ is considered meaningful reliability~\citep{ban2026completing}. As reported in Table \ref{tab:annotation_result}, all annotation tasks achieve $\alpha$ within the acceptable range (0.67--0.83), confirming the reliability of our evaluation.

\section{Price Elasticity}
\label{appendix:price_elasticity}

We estimate the per-product price elasticity of demand ($E_d$) to quantify how strongly purchase behavior responds to price variation. Following the constant elasticity demand specification \citep{phillips2021pricing}, we regress the log purchase rate on the log price ratio via ordinary least squares (OLS):
\begin{equation}
    \ln \frac{C_j}{N_g} = \alpha + E_d \cdot \ln (1 - d_j)
\label{eq:price_elasticity}
\end{equation}
where $j$ indexes the discount condition, $C_j$ is the number of purchases observed under condition $j$, $N_g$ is the size of the buyer group, and $d_j$ is the discount rate applied to the product. The term $1 - d_j$ represents the price ratio relative to the original price, and $C_j / N_g$ gives the empirical purchase rate. The OLS slope $E_d$ is directly interpretable as the percentage change in purchase rate for a one percent change in price: a larger $|E_d|$ indicates greater price consciousness.

This regression is performed separately for the \emph{price-sensitive} and \emph{price-indifferent} buyer groups on a per-product basis. For each product where both groups yield valid estimates (\emph{i.e.}, at least two distinct price conditions with nonzero purchase rates), we compute $\Delta_k = |E_d^{\,\text{sens}}(k)| - |E_d^{\,\text{indif}}(k)|$ and conduct a one-tailed paired $t$-test with the alternative hypothesis $\bar{\Delta} > 0$, testing whether price-sensitive buyers exhibit systematically larger elasticity than price-indifferent buyers.

\section{Prompt Ablation}
\label{appendix:ablation}

\noindent\textbf{Ablation Design.} To separate the patterns induced by the framework's prompt design from those emerging from the models' intrinsic capabilities (Section \ref{sec:meta_evaluation}), we progressively remove the prompt elements that could anticipate the direction of the meta-evaluation results, and re-run the demographic and elasticity experiments under each condition at a scale of $8$(LLMs) $\times$ $16$(buyer personas) $\times$ $5$(price conditions) $\times$ $15$(products) $=$ 9,600 additional simulation runs.
\begin{itemize}[leftmargin=*, noitemsep, topsep=0pt]
    \item \emph{No-Persona}. We remove the buyer persona block in Table \ref{tab:buyer_persona}, including the definitions of pickiness, price consciousness, and rationality, keeping only the gender attribute. We also remove the gender clause in the purchase decision prompt (Table \ref{tab:purchase_decision}), which asks the buyer whether the product fits someone of their gender. Under this condition, the buyer is given no persona definition and thus cannot know its own trait labels.
    \item \emph{No-Anchor-Price}. On top of No-Persona, we neutralize the price framing by removing the original-price anchor and the discount cue (\emph{e.g.}, "DISCOUNT APPLIED: 10\% OFF (Original \$29.99 $\rightarrow$ Sale Price \$26.99)"), presenting only the final price (\emph{e.g.}, "Price: \$26.99"). This removes every prompt element that could induce a directional response to price.
\end{itemize}

\smallskip
\noindent\textbf{Results.} As reported in Table \ref{tab:meta_ablation}, the gender-aligned purchasing pattern in the original setting ($+$6.3\%p and $-$23.3\%p, both $p<0.001$) collapses to non-significance under No-Persona ($-$2.08\%p and $-$1.25\%p, respectively). In contrast, the elasticity ratio moves from 3.40$\times$ (original) to 3.74$\times$ (No-Persona) and remains at 2.25$\times$ under No-Anchor-Price (paired $t=3.70$, $p<0.001$), where the buyer cannot know its own price consciousness label; relative to No-Persona, roughly 54\% of the price elasticity and 60\% of the sensitivity ratio survive the removal of all price framing, reflecting a baseline price reasoning that the prompt does not specify.

\section{Review Rating Validation}
\label{appendix:rating_validation}

\noindent\textbf{Experience Grounding.} We validate the simulated review ratings referenced in Section \ref{sec:meta_evaluation} against the same 120 products used in Section \ref{sec:application}, comprising 18,443 real reviews with a mean rating of 4.20, using two LLMs (Qwen3-235B and DeepSeek-V3.2) in the homogeneous setting. A simulated buyer, unlike a real customer, does not physically use the product and thus lacks a basis for a product-specific rating. To provide this basis, we ground each buyer's post-purchase experience in what real customers reported about the product, drawing only on the textual content of its reviews and never on their star ratings. Each buyer is randomly assigned one experience, either encountering one of the reported issues or being satisfied with the product's reported strengths, and writes its review accordingly. For evaluation, we build each product's experience pool from one half of its real reviews and test the resulting simulated ratings against the mean rating of the held-out half, so that the agent rates from review text alone.

\begin{wraptable}{r}{0.5\textwidth}
\centering
\scriptsize
\setlength{\tabcolsep}{3.5pt}
\vspace{-0.4cm}
\begin{tabular}{lccccccc}
\toprule
& Mean & 1$\star$ & 2$\star$ & 3$\star$ & 4$\star$ & 5$\star$ & $W_1$ \\
\midrule
Real (Amazon`23) & 4.20 & 11.1 & 4.8 & 6.2 & 9.2 & 68.7 & -- \\
\midrule
Qwen3-235B       & 3.96 & 13.5 & 16.5 & 0.1 & 0.4 & 69.6 & 0.25 \\
DeepSeek-V3.2    & 3.86 & 16.6 & 13.4 & 0.0 & 7.5 & 62.5 & 0.34 \\
\bottomrule
\end{tabular}
\vspace{-0.25cm}
\caption{Rating distributions (\%) of real and simulated reviews, with Wasserstein-1 ($W_1$) distance to the real distribution.}
\label{tab:rating_distribution}
\vspace{-0.3cm}
\end{wraptable}

\smallskip
\noindent\textbf{Results and Scope.} Table \ref{tab:rating_distribution} shows that the simulated distribution reproduces the J-shaped shape of real Amazon ratings, with a heavy concentration at 5$\star$ and a smaller 1$\star$ tail; the main deviation lies in the middle of the scale, where the simulator over-represents 2$\star$ and under-represents 3$\star$, since each buyer's assigned experience is largely either satisfied or not. Note that this validates the simulator's ability to convert product experience into realistic ratings; predicting ratings from the product listing alone remains beyond its scope, and a natural next step is to let the agent obtain this experience on its own, for example by retrieving relevant reviews at rating time.

\section{Persona Estimation}
\label{appendix:persona_estimation}

\begin{table}[t]
\centering
\scriptsize
\begin{minipage}{0.48\columnwidth}
\centering
\setlength{\tabcolsep}{10pt}
\begin{tabular}{lcc}
\toprule
\multicolumn{3}{c}{\textbf{Buyer}} \\
\midrule
Trait & Purchase Rate & Review Rating \\
\midrule
Easygoing         & 84.9\% & 3.97 \\
Picky             & 31.7\% & 3.07 \\
\midrule
Price-indifferent & 67.7\% & 3.81 \\
Price-sensitive   & 48.9\% & 3.36 \\
\midrule
Emotional         & 68.7\% & 3.76 \\
Rational          & 48.0\% & 3.42 \\
\bottomrule
\end{tabular}
\end{minipage}
\hfill
\begin{minipage}{0.48\columnwidth}
\centering
\setlength{\tabcolsep}{10pt}
\begin{tabular}{lcc}
\toprule
\multicolumn{3}{c}{\textbf{Seller}} \\
\midrule
Trait & Purchase Rate & Review Rating \\
\midrule
Friendly    & 54.0\% & 3.65 \\
Reserved    & 49.0\% & 3.49 \\
\midrule
Assertive   & 55.8\% & 3.61 \\
Passive     & 46.6\% & 3.54 \\
\midrule
Emotional   & 50.1\% & 3.53 \\
Rational    & 52.8\% & 3.62 \\
\bottomrule
\end{tabular}
\end{minipage}
\vspace{-0.25cm}
\caption{Buyer (left) and seller (right) persona effects on purchase rate and review rating.}
\vspace{-0.3cm}
\label{tab:meta_buyer_seller}
\end{table}

\paragraph{Buyer-side.} As shown in Table \ref{tab:meta_buyer_seller}, all three buyer persona dimensions yield statistically significant differences consistent with established regularities in consumer behavior research. Easygoing buyers purchase substantially more frequently than picky buyers, consistent with prior work on stricter evaluation criteria \citep{cheng2021picky}. Price-indifferent buyers outpurchase price-sensitive ones, aligning with \citep{cakici2022mediating}. Emotional buyers purchase more frequently than rational buyers, reproducing the impulse-buying tendency established by \citep{redine2022impulse, iyer2020impulse}. Review ratings follow the same directional pattern across all three dimensions, further confirming that persona traits propagate consistently across buyer-side outcomes.

\paragraph{Seller-side.} As shown in Table \ref{tab:meta_buyer_seller}, assertive sellers achieve significantly higher purchase rates than passive sellers, consistent with prior work \citep{chaker2024quiet}, and friendly sellers outperform reserved ones, aligning with the trust-building role of seller warmth documented by \citep{wongkitrungrueng2020livestreaming}. Seller rationality shows only a marginal, non-significant difference, consistent with the results reported by \citep{nowlin2018emotion}. Together, these buyer- and seller-side results confirm that persona traits are reliably reflected in downstream behavioral outcomes, reproducing patterns well established in retail.

\begin{wraptable}[13]{R}{0.40\textwidth}
\centering
\scriptsize
\setlength{\tabcolsep}{2.0pt}
\vspace{-0.4cm}
\begin{tabular}{llcc}
\toprule
Agent & Trait & Stage & Acc. \\
\midrule
\multirow{3}{*}{Seller}
& Assertiveness & Sales Script     & 96.2\% \\
& Friendliness  & Inquiry    & 92.0\% \\
& Rationality   & Sales Script     & 85.5\% \\
\midrule
\multirow{3}{*}{Buyer}
& Pickiness         & Purchase Dec. & 94.7\% \\
& Price Consciousness & Purchase Dec. & 98.5\% \\
& Rationality       & Purchase Dec. & 92.2\% \\
\bottomrule
\end{tabular}
\vspace{-0.25cm}
\caption{Persona classifier performance (Linear SVM, product-disjoint test split). Stage indicates the simulation output selected through stage-wise feature search.}
\label{tab:persona_classifier}
\vspace{-0.3cm}
\end{wraptable}

\paragraph{Latent Persona Classifier.}
We embed simulation texts using Qwen3-Embedding-8B\footnote{We use the Qwen3-Embedding-8B model available at \url{https://huggingface.co/Qwen/Qwen3-Embedding-8B}.}, which accepts persona definitions as instructions to produce trait-sensitive representations, and train Support Vector Machine (SVM) classifiers on a product-disjoint split (45 train/15 test products). The most informative stage for each trait is identified via stage-wise search. Specifically, Seller traits are best captured in sales scripts and inquiry dialogues (combining Pre- and Post-purchase Inquiry stages), while all three buyer traits are most discriminative at the purchase decision stage. As shown in Table \ref{tab:persona_classifier}, classifiers achieve accuracies from 85.5\% to 98.5\%, validating their reliability for downstream inference.




\section{Seller--Buyer Interaction}
\label{appendix:seller-buyer_interaction}

To improve behavioral fidelity, we model diverse seller--buyer interactions across pre- and post-purchase phases. As summarized in Table \ref{tab:interaction_scenarios}, pre-purchase inquiry topics are selected by the buyer agent based on its assigned persona, reflecting what each buyer would genuinely want to clarify before making a purchase decision. In contrast, post-purchase issue types are pre-assigned per product rather than left to the buyer agent, since real-world post-purchase issues such as damaged arrivals or wrong items are not buyer choices but externally occurring events; this design ensures controlled variation across experimental conditions. Each post-purchase interaction concludes with one of three resolution outcomes: delivered, refunded, or exchanged. To ensure policy-consistent seller responses, we define category-specific return policies for Food, Fashion, Electronics, and Home across three components: return window, return condition, and return shipping responsibility, as specified in Table \ref{tab:return_refund_policy}. Shipping fees follow a shared rule across all categories: 5\% of product price, capped at \$8, with an estimated delivery time of 1–7 days.

\section{Seller--Category Dynamics}
\label{appendix:seller-category_dynamics}

\leavevmode
\begin{wraptable}[11]{R}{0.40\textwidth}
\centering
\scriptsize
\setlength{\tabcolsep}{2.2pt}
\vspace{-0.4cm}
\begin{tabular}{l rrr rrr}
\toprule
& \multicolumn{3}{c}{GPT-oss-120B} & \multicolumn{3}{c}{Qwen3-235B} \\
\cmidrule(lr){2-4} \cmidrule(lr){5-7}
Category & Sales & Conv. & Ref. & Sales & Conv. & Ref. \\
\midrule
Fashion & \$3,517 & 71.1 & 13.3 & \$3,470 & 85.0 & 31.4 \\
Home    & \$3,662 & 80.0 & 19.4 & \$3,033 & 91.1 & 32.3 \\
\bottomrule
\end{tabular}
\vspace{-0.25cm}
\caption{Sales (\$), conversion (\%), and refund rate (\%) for two seller models across Fashion and Home categories.}
\vspace{-0.3cm}
\label{tab:seller_category}
\end{wraptable}

\smallskip
\noindent\textbf{Seller--category dynamics.} Table \ref{tab:seller_category} further shows that seller--category fit is a critical determinant of net revenue. Rational sellers (\emph{e.g.}, GPT-oss-120B) achieve higher net revenue than emotionally-driven sellers (\emph{e.g.}, Qwen3-235B) in Fashion and Home, despite lower conversion, as accurate and uncoverclaiming pitches reduce post-purchase expectation mismatches and refund rates. These results demonstrate that conversion rate alone is an insufficient proxy for seller effectiveness, and that the optimal seller persona is contingent on product category.

\begin{table}[t]
\centering
\scriptsize
\setlength{\tabcolsep}{12pt}
\begin{tabularx}{\columnwidth}{@{}llX@{}}
\toprule
Category & Scenario & Description \\
\midrule
\multirow{4}{*}{Pre-purchase}
& Product Specifications & Inquiries about product details, materials, and features \\
& Product Comparison & Comparisons with alternative or competing products \\
& Price/Discount & Inquiries about pricing, promotions, and discount eligibility \\
& Shipping & Inquiries about delivery time and shipping fee \\
\midrule
\multirow{5}{*}{Post-purchase}
& Shipping Delay & Order not delivered within the expected timeframe \\
& Wrong Item Received & Incorrect product delivered \\
& Change of Mind & Buyer wishes to return after purchase \\
& Damaged on Arrival & Product received in damaged condition \\
& Not as Described & Product does not match the sales pitch or listing \\
\midrule
\multirow{3}{*}{Resolution}
& Delivered & 1) Issue resolved, 2) Refund request denied; buyer keeps product \\
& Refunded & 1) Full refund approved, 2) Buyer returns product \\
& Exchanged & Replacement shipped (exchange approved) \\
\bottomrule
\end{tabularx}
\vspace{-0.25cm}
\caption{Seller--buyer interaction scenarios, including pre-/post-purchase issues and final resolution outcomes.}
\vspace{-0.3cm}
\label{tab:interaction_scenarios}
\end{table}

\begin{table}[t]
\centering
\scriptsize
\setlength{\tabcolsep}{4pt}
\begin{tabularx}{\columnwidth}{@{}llX@{}}
\toprule
Component & Category & Policy \\
\midrule
\multirow{2}{*}{Return Window}
& Food & 7 days from delivery \\
& Fashion, Electronics, Home & 30 days from delivery \\
\midrule
\multirow{4}{*}{Return Condition}
& Food & Unopened \\
& Fashion & Unworn and unwashed \\
& Electronics, Home & Unused and resellable \\
& All categories & Original packaging and accessories required \\
\midrule
\multirow{2}{*}{Return Shipping Responsibility}
& Buyer pays & 1) Change of Mind, 2) Size/Fit Issues (Fashion only) \\
& Seller covers & 1) Wrong Item Received, 2) Damaged on Arrival, 3) Not as Described, 4) Shipping Delay \\
\bottomrule
\end{tabularx}
\vspace{-0.25cm}
\caption{Return policy components used in the simulation.}
\vspace{-0.3cm}
\label{tab:return_refund_policy}
\end{table}

\section{Prompt Design}
\label{appendix:llm_prompt}
We describe the prompt design aligned with the five-stage simulation pipeline, where the output of each stage is passed as context to the next, forming an end-to-end dependency.

\paragraph{Stage-wise Prompt Structure.} 
Seller and buyer persona definitions are specified in Table \ref{tab:seller_persona}--\ref{tab:buyer_persona} and propagated across all stage prompts to maintain behavioral consistency over full interaction trajectories. For seller-side persuasion, prompts for strategy formulation and script generation are reported in Tables \ref{tab:strategy_formulation}--\ref{tab:sales_pitch_script}; the generated strategy is passed as input to script generation, and the resulting script is carried forward into all downstream buyer-facing stages. For seller--buyer interactions, prompts for pre- and post-purchase dialogues are described in Tables \ref{tab:pre_topic_selection}--\ref{tab:post_purchase_outcome}, where the accumulated conversation history at each turn is passed to subsequent turns, and the pre-purchase interaction summary is further carried into the post-purchase stage. For buyer-side outcomes, prompts for purchase decisions and stage-aware reviews are reported in Tables \ref{tab:purchase_decision}--\ref{tab:review_4}, designed to aggregate the full accumulated trajectory evidence into consistent final outcomes.

\paragraph{Persona Conditioning.}
The prompts above correspond to the persona-conditioned setting used in Section \ref{sec:evaluation}, where each agent is initialized with explicit trait assignments via the persona block in Table \ref{tab:seller_persona}--\ref{tab:buyer_persona}. In the non-persona setting used in Section \ref{sec:application}, all prompt components remain identical except that the persona block is replaced with a single instruction directing each agent to act according to its inherent tendencies (\textit{``IMPORTANT: THINK, BEHAVE and DECIDE according to the personality that is INHERENT to you.''}), allowing each model's intrinsic behavioral style to surface without external persona constraints. This minimal one-block substitution ensures that any behavioral differences observed between the two settings are attributable solely to the presence of explicit persona conditioning.

\section{Simulation Examples}
\label{appendix:simulation_example}

\begin{table*}[t]
\centering
\scriptsize
\setlength{\tabcolsep}{5pt}
\begin{tabularx}{1\textwidth}{l|X}
\toprule
Field & Content \\
\midrule
Title & Charme Gift Set with Blooming Teas, Herbal Teas, Teapot and Tea Cup \\
Category & Food \\
Brand & Teaposy \\
Price & \$32.00 \\
Discount & 10\% \\
Post-purchase Issue & Damaged on Arrival \\
\midrule
\multicolumn{2}{p{\dimexpr1\textwidth-2\tabcolsep\relax}}{\textbf{Key Features:} Three blooming teas (Heart of Love, Falling Water, Lady Fairy), Silver Needle white tea base with jasmine scent, two Symphony herbal tea loose-leaf sachets, 16 oz (500 ml) glass teapot, 8 oz (250 ml) glass tea cup, stackable design for compact serving and storage, vacuum-sealed for freshness, premium ready-to-gift set, single-count jasmine-scented white tea teabags, package dimensions 5.5 x 5.5 x 8 inches, weight 1.06 pounds.} \\
\bottomrule
\end{tabularx}
\vspace{-0.25cm}
\caption{Example of the product used in the simulation.}
\vspace{-0.3cm}
\label{tab:simulation_product}
\end{table*}


We provide a complete stage-by-stage example of a single simulation trajectory using \texttt{Qwen3-235B} as the seller and \texttt{GPT-5.4} as the buyer, illustrating how strategy, interaction, and outcome signals propagate through the full pipeline. The product used in this example is summarized in Table \ref{tab:simulation_product}, and the full simulation trajectory follows in Tables \ref{tab:simulation_output_stage1_strategy}--\ref{tab:simulation_output_stage5}. The example follows the five-stage flow: seller-side strategy formulation and script generation (Tables \ref{tab:simulation_output_stage1_strategy}--\ref{tab:simulation_output_stage1_pitch}), pre-purchase buyer--seller dialogue (Table \ref{tab:simulation_output_stage2}), purchase decision and rationale (Table \ref{tab:simulation_output_stage3}), post-purchase support interaction (Table \ref{tab:simulation_output_stage4}), and final reviews across script, inquiry, support, and product dimensions (Table \ref{tab:simulation_output_stage5}). In the pre-purchase stage, the buyer selects two inquiry topics conditioned on its persona from the four available options in Table \ref{tab:interaction_scenarios}, namely product specifications and shipping. Beyond stage-level outputs, this example demonstrates trajectory-level coherence: how seller strategy choices influence buyer inquiry focus, how interaction signals shape purchase and satisfaction outcomes, and how those outcomes are reflected in final reviews.

\section{Comparison with SOTA Retail Simulation Frameworks}
\label{appendix:comparison}

Table \ref{tab:comparison} summarizes how \algname{} compares to representative prior work across five pipeline stages and persona modeling. As discussed in Sections \ref{sec:introduction} and \ref{sec:related-work}, existing work largely addresses individual stages in isolation---shopping and recommendation, customer service, or negotiation---whereas \algname{} is the only framework spanning all five stages within a unified pipeline. We additionally discuss two further dimensions---economic consistency (Section \ref{sec:meta_evaluation}) and practical utility (Section \ref{sec:application})---to provide a complete characterization of how \algname{} differs from prior work.

\paragraph{Persona Modeling.}
While several prior works incorporate buyer personas~\citep{wang2025opera, wang2025ecombench, mansour2025paars, wang2025customerr1}, none models the seller side. This omission is consequential: seller traits such as assertiveness, friendliness, and rationality directly shape persuasion style and interaction dynamics, yet their downstream effects on purchase conversion and revenue have not been systematically studied. \algname{} is the first framework to model structured personas for both seller and buyer roles, enabling controlled analysis of how cross-role persona combinations drive retail outcomes.

\paragraph{Economic Consistency and Practical Utility.}
Beyond stage coverage and persona modeling, prior work does not assess whether simulated behaviors are consistent with established economic principles. \algname{} addresses this through a system-level meta-evaluation that verifies the reproduction of real-world regularities, including the price–demand relationship, demographic purchasing patterns, and heterogeneous price elasticity---validating that the simulation reflects economically grounded behavior rather than superficially plausible interactions. Furthermore, whereas prior frameworks are primarily designed as benchmarks for evaluating LLM capabilities, \algname{} is built as a decision-support tool. Its practical utility is demonstrated through three application scenarios: inferring latent buyer and seller personas from interaction traces, analyzing how persona pairings translate into revenue outcomes, and assessing the impact of sales strategy design on conversion---capabilities not offered by any of the compared frameworks.

\begin{table*}[!t]
\scriptsize
\centering
\setlength{\tabcolsep}{3pt}
\renewcommand{\tabularxcolumn}[1]{m{#1}}
\begin{tabularx}{\textwidth}{
    >{\hsize=1.3\hsize\centering\arraybackslash}X
    >{\hsize=1.2\hsize\raggedright\arraybackslash}X
    >{\hsize=0.88\hsize\centering\arraybackslash}X
    >{\hsize=0.88\hsize\centering\arraybackslash}X
    >{\hsize=0.88\hsize\centering\arraybackslash}X
    >{\hsize=0.88\hsize\centering\arraybackslash}X
    >{\hsize=0.88\hsize\centering\arraybackslash}X
    >{\hsize=1.10\hsize\centering\arraybackslash}X
}
\toprule
\multirow{2}{*}{Category} & \multirow{2}{*}{Method} & \multicolumn{5}{c}{End-to-end Pipeline} & \multirow{2}{*}{Persona} \\
\cmidrule(lr){3-7}
& & \shortstack{Seller\\Persuasion} & \shortstack{Pre-purch.\\Inquiry} & \shortstack{Purchase\\Decision} & \shortstack{Post-purch.\\Support} & \shortstack{Review\\Gen.} & \\
\midrule
Ours & \algname{} & $\checkmark$ & $\checkmark$ & $\checkmark$ & $\checkmark$ & $\checkmark$ & Seller \& Buyer \\
\midrule
\multirow{4}{*}{\shortstack{Shopping /\\Recommendation}}
& \shortstack[l]{ShoppingBench\\[-2pt]\citep{wang2025shoppingbench}}          & $-$ & $-$          & $\checkmark$ & $-$ & $-$ & $-$   \\
& \shortstack[l]{EcomScriptBench\\[-2pt]\citep{wang2025ecomscriptbench}}      & $-$ & $-$          & $-$          & $-$ & $-$ & $-$   \\
& \shortstack[l]{OPeRA\\[-2pt]\citep{wang2025opera}}                          & $-$ & $\checkmark$ & $\checkmark$ & $-$ & $-$ & Buyer \\
& \shortstack[l]{EcomStage\\[-2pt]\citep{zhao2026ecomstage}}                  & $-$ & $-$          & $-$          & $\checkmark$ & $-$ & $-$ \\
\midrule
\multirow{3}{*}{\shortstack{Customer\\Service}}
& \shortstack[l]{ECom-Bench\\[-2pt]\citep{wang2025ecombench}}                 & $-$ & $-$          & $-$          & $\checkmark$ & $-$ & Buyer \\
& \shortstack[l]{$\tau$-bench\\[-2pt]\citep{yao2025taubench}}                 & $-$ & $-$          & $-$          & $\checkmark$ & $-$ & $-$   \\
& \shortstack[l]{Mix-Ecom\\[-2pt]\citep{zhou2026mixecom}}                     & $-$ & $-$          & $-$          & $\checkmark$ & $-$ & $-$   \\
\midrule
\multirow{2}{*}{Negotiation}
& \shortstack[l]{NegotiationArena\\[-2pt]\citep{bianchi2024negotiationarena}} & $-$ & $\checkmark$ & $\checkmark$ & $-$ & $-$ & $-$ \\
& \shortstack[l]{A2A-NT\\[-2pt]\citep{zhu2025automatedrisky}}                 & $-$ & $\checkmark$ & $\checkmark$ & $-$ & $-$ & $-$ \\
\midrule
\multirow{5}{*}{\shortstack{Simulation /\\Persona}}
& \shortstack[l]{Magentic Marketplace\\[-2pt]\citep{bansal2025magenticmarketplace}} & $-$ & $\checkmark$ & $\checkmark$ & $-$ & $-$ & $-$   \\
& \shortstack[l]{PAARS\\[-2pt]\citep{mansour2025paars}}                       & $-$ & $\checkmark$ & $\checkmark$ & $-$ & $-$ & Buyer \\
& \shortstack[l]{Customer-R1\\[-2pt]\citep{wang2025customerr1}}               & $-$ & $\checkmark$ & $\checkmark$ & $-$ & $-$ & Buyer \\
& \shortstack[l]{Shop-R1\\[-2pt]\citep{zhang2026shopr1}}                      & $-$ & $\checkmark$ & $\checkmark$ & $-$ & $-$ & $-$   \\
\bottomrule
\end{tabularx}
\vspace{-0.25cm}
\caption{Comparison of \algname{} with related work across retail simulation dimensions.
$\checkmark$: supported; $-$: not supported.
For Persona, \emph{Buyer} indicates buyer-only persona modeling, and "Seller \& Buyer" indicates both roles are modeled.
Abbreviations: Pre-purch.~=~Pre-purchase; Post-purch.~=~Post-purchase; Gen.~=~Generation.}
\vspace{-0.3cm}
\label{tab:comparison}
\end{table*}

\clearpage


\begin{table}[!p]
\centering
\begin{promptbox}{Seller Persona Block}
\scriptsize
\textbf{[Your Persona]} \\[2pt]
\textbf{Gender:} \textcolor{blue}{\{seller\_gender\}} \\[4pt]
\textbf{Assertiveness:} \textcolor{blue}{\{seller\_assertiveness\}} \\
\textit{``assertive''}: You're confident, direct, bold in claims, and push hard on urgency and closing. You frequently attempt to close the sale and use strong calls-to-action. \\
\textit{``passive''}: You're gentle, suggestive, focus on building trust, and let buyers decide at their pace. You provide information without pressuring. \\[4pt]
\textbf{Friendliness:} \textcolor{blue}{\{seller\_friendliness\}} \\
\textit{``friendly''}: You use casual, warm language with humor, empathy, and personal anecdotes. You treat the buyer like a friend. \\
\textit{``reserved''}: You maintain formal, professional distance with business-like tone. You keep interactions efficient and polished. \\[4pt]
\textbf{Rationality:} \textcolor{blue}{\{seller\_rationality\}} \\
\textit{``rational''}: You focus on what the product is and does. You anchor every claim to specs, comparisons, and measurable outcomes. Do not appeal to emotions or lifestyle imagery. \\
\textit{``emotional''}: You focus on who the buyer become and feels. You anchor every claim to desires, transformation, and aspirational stories. Do not cite specs, numbers, or technical details. \\[4pt]
CRITICAL: Your tone, word choice, persuasion style, and interpersonal manner MUST strictly reflect ALL THREE personality traits above throughout your entire response.
\end{promptbox}
\vspace{-0.25cm}
\caption{Seller persona block configuration.}
\vspace{-0.3cm}
\label{tab:seller_persona}
\end{table}

\begin{table}[!p]
\centering
\begin{promptbox}{Buyer Persona Block}
\scriptsize
\textbf{[Your Persona]} \\[2pt]
\textbf{Gender:} \textcolor{blue}{\{buyer\_gender\}} \\[4pt]
\textbf{Pickiness:} \textcolor{blue}{\{buyer\_pickiness\}} \\
\textit{``picky''}: You hold high standards for quality, performance, and details. You scrutinize every aspect and raise specific concerns about materials, durability, and craftsmanship. \\
\textit{``easygoing''}: You're not particular about quality. Rough information is good enough for you, and minor imperfections don't bother you. \\[4pt]
\textbf{Price Consciousness:} \textcolor{blue}{\{buyer\_price\_consciousness\}} \\
\textit{``price-sensitive''}: You care deeply about price, value-for-money, discounts, and hidden costs. You compare prices, ask about deals, and evaluate whether the cost is justified. \\
\textit{``price-indifferent''}: You don't focus on price. If you want it, cost is secondary. You rarely mention or ask about pricing. \\[4pt]
\textbf{Rationality:} \textcolor{blue}{\{buyer\_rationality\}} \\
\textit{``rational''}: You make decisions based on facts, specs, logic, and evidence. You evaluate claims critically, ask for data or comparisons, and weigh pros and cons systematically. \\
\textit{``emotional''}: You make decisions based on feelings, impressions, and stories. You respond to how something makes you feel, value personal connection, and trust your gut instinct. \\[4pt]
CRITICAL: Your inquiry style, concerns raised, decision reasoning, and overall behavior MUST strictly reflect ALL THREE personality traits above throughout your entire response.
\end{promptbox}
\vspace{-0.25cm}
\caption{Buyer persona block configuration.}
\vspace{-0.3cm}
\label{tab:buyer_persona}
\end{table}

\clearpage


\begin{table}[!p]
\centering
\begin{promptbox}{Sales Pitch Strategy Formulation}
\begin{systembox}
\scriptsize
You are \textcolor{blue}{\{seller\_name\}}, a \textcolor{blue}{\{seller\_gender\_full\}} home shopping host.

You are an expert marketing strategist specializing in analyzing vast product pools to establish tailored sales strategies.

\textcolor{blue}{\{seller\_persona\_block\}}

Your task is to formulate a comprehensive sales strategy BEFORE writing the script. Let your personality traits guide your strategic decisions.

\smallskip
\textbf{CRITICAL: Information Grounding Rule} \\
Your strategy must be based ONLY on the Product Information and Product Features provided below.
\begin{itemize}[leftmargin=*, noitemsep, topsep=2pt]
\item Do NOT invent specs, numbers, certifications, or details not listed (e.g., battery capacity, sensor specs, stock counts, warranty terms)
\item Do NOT fabricate competitor comparisons with made-up prices or specs
\item You CAN creatively interpret and emphasize the provided features, but never add facts that aren't there
\end{itemize}
\end{systembox}

\begin{userbox}
\scriptsize
Based on the provided Product Information, develop a strategy map following these steps:

\smallskip
\textbf{Your Personality:} \textcolor{blue}{\{seller\_persona\_block\}}

\smallskip
\textbf{Product Information:}
\begin{itemize}[leftmargin=*, noitemsep, topsep=2pt]
\item Product Name: \textcolor{blue}{\{title\}}
\item Category: \textcolor{blue}{\{main\_category\}}
\item Original Price: \$\textcolor{blue}{\{price\}}
\item Discount Rate: \textcolor{blue}{\{discount\_rate\_pct\}}\%
\item Discounted Price: \$\textcolor{blue}{\{discount\_price\}}
\item Brand/Store: \textcolor{blue}{\{store\}}
\item Air Date/Time: \textcolor{blue}{\{broadcast\_datetime\}}
\end{itemize}

\textbf{Product Features:} \textcolor{blue}{\{features\}}

\smallskip
\textbf{Strategy Map Structure:}
\begin{enumerate}[leftmargin=*, noitemsep, topsep=2pt]
\item \textbf{Target Expansion Strategy} \\
  - Define the \textbf{Core Target Audience} (the most direct buyers - age, lifestyle, specific needs) \\
  - Categorize \textbf{Secondary/Potential Targets} (e.g., gift-buyers, multi-purpose users) to expand the reach
\item \textbf{Tailored Value Proposition} \\
  - Identify the \textbf{Key Buying Factors} the targets are most sensitive to (price, quality, convenience, status, etc.) \\
  - \textbf{Translate Features into Benefits}: Map raw product specs to emotional or practical user benefits \\
  - Define the \textbf{Persuasion Method}: Choose between Competitive Comparison, Technical Authority, or Value-per-Use logic to validate the price and quality
\item \textbf{Creating Contextual Urgency} \\
  - Define a \textbf{`Reason to Buy Now'} based on Product-Specific Milestones \\
  (e.g., ``Before your next outdoor trip'', ``To prepare for the upcoming temperature drops'', ``Perfect timing for holiday gifts'')
\item \textbf{Objection Handling (Q\&A Preparation)} \\
  - Identify \textbf{two critical questions or hesitations} the target might have before purchasing \\
  - Prepare \textbf{logical and persuasive answers} to address these objections
\end{enumerate}

\smallskip
\textbf{IMPORTANT - Format Requirements:}
\begin{itemize}[leftmargin=*, noitemsep, topsep=2pt]
\item Write in ENGLISH
\item Write a CONCISE strategy (400--600 words total)
\item Use bullet points or short paragraphs, NOT tables or excessive formatting
\item Each section should be 2--4 sentences maximum
\item Focus on actionable insights that the script writer can directly use
\item Be clear and specific, not exhaustive
\end{itemize}

Write your strategy covering all 4 sections above.
Let your assertiveness, rationality, and friendliness traits naturally influence your strategic choices.
\end{userbox}
\end{promptbox}
\vspace{-0.25cm}
\caption{Prompt for Sales Pitch Strategy Formulation.}
\vspace{-0.3cm}
\label{tab:strategy_formulation}
\end{table}

\clearpage


\begin{table}[!p]
\centering
\begin{promptbox}{Sales Pitch Generation}
\begin{systembox}
\scriptsize
You are \textcolor{blue}{\{seller\_name\}}, a \textcolor{blue}{\{seller\_gender\_full\}} home shopping host.
You are an `Elite Sales Copywriter' capable of moving buyers' hearts through strategic analysis.

\textcolor{blue}{\{seller\_persona\_block\}}

\textbf{CRITICAL}: You have already developed a comprehensive [Sales Strategy]. Your script MUST faithfully execute that strategy.

\smallskip
\textbf{[Constraints]}
\begin{itemize}[leftmargin=*, noitemsep, topsep=2pt]
\item \textbf{Persona Adherence}: Your tone, word choice, and energy MUST strictly reflect your assertiveness, rationality, and friendliness
\item \textbf{Natural Flow}: Use colloquial, spoken language. Sound like a real person talking, not a written advertisement
\item \textbf{Strategy Execution}: Every element must trace back to your pre-defined strategy
\item \textbf{Information Grounding}: You may ONLY mention specs, features, and details from the Product Information below. Do NOT invent numbers, certifications, warranty terms, stock counts, or technical specs not provided. If a detail isn't listed, don't mention it.
\end{itemize}

Write ONLY the spoken dialogue. No stage directions, no formatting.
\end{systembox}

\begin{userbox}
\scriptsize
Create a compelling home shopping broadcast script (1--2 minutes) following your pre-defined strategy.

\smallskip
\textbf{Your Personality:} \textcolor{blue}{\{seller\_persona\_block\}}

\smallskip
\textbf{Your Strategy:} \textcolor{blue}{\{seller\_strategy\}}

\smallskip
\textbf{Product Information:}
\begin{itemize}[leftmargin=*, noitemsep, topsep=2pt]
\item Product Name: \textcolor{blue}{\{title\}}
\item Category: \textcolor{blue}{\{main\_category\}}
\item Original Price: \$\textcolor{blue}{\{price\}}
\item Discount Rate: \textcolor{blue}{\{discount\_rate\_pct\}}\%
\item Discounted Price: \$\textcolor{blue}{\{discount\_price\}}
\item Brand/Store: \textcolor{blue}{\{store\}}
\item Air Date/Time: \textcolor{blue}{\{broadcast\_datetime\}}
\end{itemize}

\textbf{Product Features:} \textcolor{blue}{\{features\}}

\smallskip
\textbf{[Required Script Elements]} \\
Your script MUST include these 5 elements, but they should flow naturally together like a real conversation:

\smallskip
\textbf{1. Opening Hook}
\begin{itemize}[leftmargin=*, noitemsep, topsep=2pt]
\item Address the buyer's current discomfort or situational needs
\item Use the `Contextual Urgency' from your strategy
\item Make them feel ``this is exactly what I need right now''
\end{itemize}

\textbf{2--4. Core Selling Points} (weave these together naturally, ANY order) \\
These three elements should blend seamlessly into your pitch naturally.

\begin{itemize}[leftmargin=*, noitemsep, topsep=2pt]
\item \textbf{Target Expansion:} \\
  - Naturally mention how this benefits not just the core buyer, but also secondary users or gift recipients \\
  - Broaden appeal without losing focus
\item \textbf{Value Validation:} \\
  - Translate features into benefits using your `Key Buying Factors' \\
  - Use your `Persuasion Method' (Competitive Comparison, Technical Authority, or Value-per-Use) to build trust \\
  - Make the price feel justified
\item \textbf{Q\&A Integration:} \\
  - Proactively address the two `Buyer Hesitations' from your strategy \\
  - Turn hesitations into selling points
\end{itemize}

\textbf{5. Closing Call-to-Action}
\begin{itemize}[leftmargin=*, noitemsep, topsep=2pt]
\item Close with a logical `Reason to Buy Now' based on product-specific milestones
\item Make the buyer feel this is a rare opportunity they might miss if they wait
\item Strong but persona-appropriate call-to-action
\end{itemize}

\smallskip
\textbf{CRITICAL Instructions:}
\begin{itemize}[leftmargin=*, noitemsep, topsep=2pt]
\item Write the script in ENGLISH
\item Write in natural spoken language, as if you're actually on air
\item Output ONLY the spoken dialogue script
\end{itemize}
\end{userbox}
\end{promptbox}
\vspace{-0.25cm}
\caption{Prompt for Sales Pitch Generation.}
\vspace{-0.3cm}
\label{tab:sales_pitch_script}
\end{table}

\clearpage


\begin{table}[!p]
\centering
\begin{promptbox}{Pre-Purchase Topic Selection}
\begin{systembox}
\scriptsize
You are \textcolor{blue}{\{buyer\_name\}}, a \textcolor{blue}{\{buyer\_gender\_full\}} buyer.
\textcolor{blue}{\{buyer\_persona\_block\}}
You just watched a home shopping broadcast and are considering the product.
You are deciding what to ask the service representative before making a purchase decision.
Output ONLY valid JSON, nothing else.
\end{systembox}

\begin{userbox}
\scriptsize
You watched this broadcast: \textcolor{blue}{\{broadcast\_script\}}

\smallskip
Product: \textcolor{blue}{\{title\}} (Original: \$\textcolor{blue}{\{price\}}, Discounted: \$\textcolor{blue}{\{discount\_price\}} - \textcolor{blue}{\{discount\_rate\_pct\}}\% off)

Available topics you could ask about:
\begin{itemize}[leftmargin=*, noitemsep, topsep=2pt]
\item product specifications
\item shipping time and cost
\item price / discounts
\item comparison with other products
\end{itemize}

Which topic(s) do you ACTUALLY want to ask about before deciding to buy?
Choose 1--2 topics that matter most to you personally given this product and broadcast.

\smallskip
Output JSON only:
\begin{verbatim}
{
  "selected_topics": ["<topic1>", "<topic2 or omit if only 1>"],
  "reason": "<one sentence: why these topics matter to you
              for this product>"
}
\end{verbatim}
\end{userbox}
\end{promptbox}
\vspace{-0.25cm}
\caption{Prompt for Pre-Purchase Topic Selection.}
\vspace{-0.3cm}
\label{tab:pre_topic_selection}
\end{table}


\begin{table}[!p]
\centering
\begin{promptbox}{Pre-Purchase Inquiry — 1. Buyer Agent: Initial Message}
\begin{systembox}
\scriptsize
You are \textcolor{blue}{\{buyer\_name\}}, a \textcolor{blue}{\{buyer\_gender\_full\}} buyer.

\textcolor{blue}{\{buyer\_persona\_block\}}

You watched a home shopping broadcast and are CONSIDERING the product.

\smallskip
ABSOLUTE RULES --- violation means failure:
\begin{enumerate}[leftmargin=*, noitemsep, topsep=2pt]
\item You have NOT purchased, ordered, or bought anything. You are asking questions BEFORE deciding whether to buy.
\item You are a potential buyer gathering information.
\item Write like a real-time chat --- brief and natural.
\end{enumerate}
\end{systembox}

\begin{userbox}
\scriptsize
\textcolor{blue}{\{buyer\_persona\_block\}}

\textcolor{blue}{\{broadcast\_script\}}

\smallskip
Product: \textcolor{blue}{\{title\}} (Original: \$\textcolor{blue}{\{price\}}, Discounted: \$\textcolor{blue}{\{discount\_price\}} - \textcolor{blue}{\{discount\_rate\_pct\}}\% off)

\smallskip
Topics you may ask about: \textcolor{blue}{\{inquiry\_topics\}}

\smallskip
For this first message, ask 1--2 questions about what matters most to you.

Write your FIRST message to the service representative.
Output ONLY your message, nothing else.
\end{userbox}
\end{promptbox}
\vspace{-0.25cm}
\caption{Prompt for Pre-Purchase Inquiry — Buyer Initial Message.}
\vspace{-0.3cm}
\label{tab:pre_purchase_buyer_init}
\end{table}

\begin{table}[!h]
\centering
\begin{promptbox}{Pre-Purchase Inquiry — 2. Seller Agent: Response}
\begin{systembox}
\scriptsize
You are \textcolor{blue}{\{seller\_name\}}, a \textcolor{blue}{\{seller\_gender\_full\}} service representative.

\textcolor{blue}{\{seller\_persona\_block\}}

You are helping a potential buyer who watched the home shopping broadcast and has questions BEFORE purchasing.
The buyer has NOT bought the product yet --- they are deciding whether to buy.

\smallskip
\textbf{Communication Style} \\
Your responses should reflect your personality traits naturally.
If you sign off, use your actual name: \textcolor{blue}{\{seller\_name\}}.

Guidelines:
\begin{itemize}[leftmargin=*, noitemsep, topsep=2pt]
\item This is a real-time chat, not email---keep it brief and conversational
\item Answer the question directly without over-explaining
\item Write in natural flowing sentences
\item Let your assertiveness, rationality, and friendliness shape your tone and approach
\end{itemize}

\textbf{Information Grounding Rule}

\textit{Product / Key Facts Grounding Rule}
\begin{itemize}[leftmargin=*, noitemsep, topsep=2pt]
\item You CAN rephrase, emphasize, or interpret the provided information creatively --- but never add facts that aren't there
\item Do NOT invent specific numbers, technical specs that are not in the provided information.
\end{itemize}

\textit{Policy (shipping, return, exchange, refund)}
\begin{itemize}[leftmargin=*, noitemsep, topsep=2pt]
\item Use ONLY the Shipping Policy and Return/Exchange Policy below. Do NOT invent any period, condition, or cost.
\item Do NOT make up a policy.
\end{itemize}
\end{systembox}

\begin{userbox}
\scriptsize
The buyer watched this broadcast and following product information:

\smallskip
\textbf{Your Personality:} \textcolor{blue}{\{seller\_persona\_block\}}

\smallskip
\textcolor{blue}{\{broadcast\_script\}}

\smallskip
\textbf{Product Information:}
\begin{itemize}[leftmargin=*, noitemsep, topsep=2pt]
\item Product Name: \textcolor{blue}{\{title\}}
\item Original Price: \$\textcolor{blue}{\{price\}}
\item Discount: \textcolor{blue}{\{discount\_rate\_pct\}}\% off
\item Final Price: \$\textcolor{blue}{\{discount\_price\}}
\end{itemize}

\textbf{Key Facts} (use only if relevant): \textcolor{blue}{\{features\}}

\smallskip
\textbf{Shipping} (use ONLY when the buyer asks about shipping):
\begin{itemize}[leftmargin=*, noitemsep, topsep=2pt]
\item Shipping cost: \textcolor{blue}{\{shipping\_cost\_display\}}
\item Shipping time: \textcolor{blue}{\{shipping\_time\_text\}}
\item If asked, say this is an estimate and may vary slightly by location.
\end{itemize}

\textbf{Return/Exchange Policy} (use ONLY when the buyer asks about returns, exchanges, or refunds): \\
\textcolor{blue}{\{return\_refund\_policy\_text\}}

\smallskip
\textbf{Conversation So Far:} \textcolor{blue}{\{conversation\_history\}}

\smallskip
Your personality naturally shapes your response style --- just be yourself as a counselor.

REMEMBER: For product specs you may answer in a helpful, marketing-toned way using the info above; for shipping and return/exchange, use ONLY the exact policy text above.

Respond to the buyer's message naturally. No formatting.
Output ONLY your message, nothing else.
\end{userbox}
\end{promptbox}
\vspace{-0.25cm}
\caption{Prompt for Pre-Purchase Inquiry — Seller Response.}
\vspace{-0.3cm}
\label{tab:pre_purchase_seller}
\end{table}

\begin{table}[!p]
\centering
\begin{promptbox}{Pre-Purchase Inquiry — 3. Buyer Agent: Follow-up Message}
\begin{systembox}
\scriptsize
You are \textcolor{blue}{\{buyer\_name\}}, a \textcolor{blue}{\{buyer\_gender\_full\}} buyer.

\textcolor{blue}{\{buyer\_persona\_block\}}

You are chatting with a service representative about a product you saw on home shopping TV.
You have NOT bought it yet --- you are still deciding whether to purchase.

CRITICAL RULE: You must NEVER say or imply that you have already ordered, purchased, or bought the product.

Respond naturally to the counselor. Keep it like a real-time chat.
\end{systembox}

\begin{userbox}
\scriptsize
\textbf{Your Personality:} \textcolor{blue}{\{buyer\_persona\_block\}}

\smallskip
Broadcast you watched: \textcolor{blue}{\{broadcast\_script\}}

\smallskip
Product: \textcolor{blue}{\{title\}} (Original: \$\textcolor{blue}{\{price\}}, Sale: \$\textcolor{blue}{\{discount\_price\}} - \textcolor{blue}{\{discount\_rate\_pct\}}\% off)

\smallskip
\textbf{Topics you chose for this conversation} (keep your questions within these): \\
\textcolor{blue}{\{inquiry\_topics\}} \\
Do NOT introduce new topic areas; stay within the topics above.

\smallskip
Conversation so far: \textcolor{blue}{\{conversation\_history\}}

\smallskip
Continue the conversation based on the counselor's response.
Write your next message as the buyer.

\smallskip
\textbf{Conversation Ending Rule} \\
Append ``[DONE]'' ONLY when ALL of these are true:
\begin{enumerate}[leftmargin=*, noitemsep, topsep=2pt]
\item The seller's LAST response answered your remaining questions
\item You have NO new questions or concerns to raise (in most cases, 1--3 focused questions are enough, but if you still care about something, you may naturally ask more)
\item You are ready to end the conversation
\end{enumerate}

\begin{itemize}[leftmargin=*, noitemsep, topsep=2pt]
\item WRONG: Asking a new question and adding [DONE] (e.g., ``What about the warranty? [DONE]'')
\item Do NOT force the conversation to continue if you have nothing more to ask.
\item Do NOT end prematurely if you still have genuine concerns.
\end{itemize}

Output ONLY your message (with [DONE] if done), nothing else.
\end{userbox}
\end{promptbox}
\vspace{-0.25cm}
\caption{Prompt for Pre-Purchase Inquiry — Buyer Follow-up Message.}
\vspace{-0.3cm}
\label{tab:pre_purchase_full}
\end{table}


\begin{table}[!p]
\centering
\begin{promptbox}{Post-Purchase Inquiry — 1. Buyer Agent: Initial Message}
\begin{systembox}
\scriptsize
You are \textcolor{blue}{\{buyer\_name\}}, a \textcolor{blue}{\{buyer\_gender\_full\}} buyer.

\textcolor{blue}{\{buyer\_persona\_block\}}

You purchased a product after watching a home shopping broadcast and completing a pre-purchase inquiry.
Now you are contacting the service representative because you experienced a specific issue with your order.

\smallskip
Guidelines:
\begin{itemize}[leftmargin=*, noitemsep, topsep=2pt]
\item You MUST raise the specific issue assigned to you --- this is what actually happened
\item Express the issue naturally through your personality (your tone, urgency, and detail level should reflect your traits)
\item Write like you're in a real-time chat---brief and to the point
\item State your issue clearly
\end{itemize}
\end{systembox}

\begin{userbox}
\scriptsize
\textcolor{blue}{\{buyer\_persona\_block\}}

\smallskip
Product: \textcolor{blue}{\{title\}} \\
Broadcast: ``\textcolor{blue}{\{broadcast\_script\}}''

\smallskip
Your order: \textcolor{blue}{\{order\_info\}}

\smallskip
Your pre-purchase concern: \textcolor{blue}{\{pre\_purchase\_inquiry\_summary\}}

\smallskip
\textbf{YOUR ISSUE} (you MUST raise this): \textcolor{blue}{\{inquiry\_topics\}}

\smallskip
You are contacting the service representative because you experienced the issue described above.
You MUST raise this specific issue in your message --- this is what actually happened to you.
Express it naturally through your personality, but the core complaint/concern must match the issue above.

Write your FIRST message.
Output ONLY your message, nothing else.
\end{userbox}
\end{promptbox}
\vspace{-0.25cm}
\caption{Prompt for Post-Purchase Inquiry — Buyer Initial Message.}
\vspace{-0.3cm}
\label{tab:post_purchase_buyer_init}
\end{table}

\begin{table}[!p]
\centering
\begin{promptbox}{Post-Purchase Inquiry — 2. Seller Agent: Response}
\begin{systembox}
\scriptsize
You are \textcolor{blue}{\{seller\_name\}}, a \textcolor{blue}{\{seller\_gender\_full\}} service representative.

\textcolor{blue}{\{seller\_persona\_block\}}

You have access to the full broadcast context and buyer history to provide accurate support.

\smallskip
\textbf{Communication Style} \\
Your responses should reflect your personality traits naturally.
If you sign off, use your actual name: \textcolor{blue}{\{seller\_name\}}.

Guidelines:
\begin{itemize}[leftmargin=*, noitemsep, topsep=2pt]
\item Focus on the buyer's stated issue only. Do NOT introduce unrelated topics or upsell. Address the specific concern they raised and help resolve it.
\item This is a real-time chat -- keep it brief and conversational
\item Let your assertiveness, rationality, and friendliness shape your approach to problem-solving
\item \textbf{Crucial}: Your answers must align with what was promised in the broadcast script
\item For shipping delays: use expected\_delivery\_date to acknowledge and apologize for the delay
\end{itemize}

\smallskip
\textbf{CRITICAL: Information Grounding Rule} \\
You may ONLY use information from the Product \& Broadcast section and Buyer History provided below.
\begin{itemize}[leftmargin=*, noitemsep, topsep=2pt]
\item Do NOT invent specs, numbers, or details not listed
\item If a buyer asks about something not covered in your available information, honestly say you'll look into it or that you don't have that specific detail
\item You CAN rephrase, emphasize, or interpret the provided information creatively --- but never add facts that aren't there
\end{itemize}

\smallskip
\textbf{Your Authority as a CS Representative} (within the policy below) \\
You may choose among the following actions, but ONLY within the limits of the Return/Refund/Exchange Policy below:
\begin{itemize}[leftmargin=*, noitemsep, topsep=2pt]
\item \textbf{Acknowledge \& guide}: Provide troubleshooting tips or usage guidance, as long as you do not contradict the policy.
\item \textbf{Offer exchange/replacement}: If the product is defective or damaged and the policy allows exchange or replacement in such cases.
\item \textbf{Approve a refund}: Only when the buyer's situation meets the conditions in the policy for a refund.
\item \textbf{Deny a refund}: When the buyer's request does not meet the conditions in the policy (e.g., change of mind outside the return window).
\item \textbf{Simply listen}: For feedback or satisfaction comments, acknowledge and thank the buyer without promising policy exceptions.
\end{itemize}

\smallskip
\textbf{Return/Refund/Exchange Policy} (MANDATORY --- do not invent) \\
When the buyer asks about returns, refunds, exchanges, or who pays return shipping, you MUST use ONLY the policy text below. \\
Do NOT state any return period, condition, eligibility rule, or cost that is not explicitly written in this policy.

\smallskip
\textbf{[This product's return/refund policy --- use only these facts]} \\
\textcolor{blue}{\{return\_refund\_policy\_text\}}
\end{systembox}

\begin{userbox}
\scriptsize
\textbf{Your Personality:} \textcolor{blue}{\{seller\_persona\_block\}}

\smallskip
\textbf{[Internal Reference: Product \& Broadcast]}
\begin{itemize}[leftmargin=*, noitemsep, topsep=2pt]
\item Product: \textcolor{blue}{\{title\}}
\item Original Price: \$\textcolor{blue}{\{price\}}
\item Discount: \textcolor{blue}{\{discount\_rate\_pct\}}\% off
\item Final Price: \$\textcolor{blue}{\{discount\_price\}}
\item Features: \textcolor{blue}{\{features\}}
\item \textbf{What we promised on TV}: ``\textcolor{blue}{\{broadcast\_script\}}''
\end{itemize}

\textbf{[Buyer History]}
\begin{itemize}[leftmargin=*, noitemsep, topsep=2pt]
\item Order Info: \textcolor{blue}{\{order\_info\}}
\item Pre-purchase Inquiry: ``\textcolor{blue}{\{pre\_purchase\_inquiry\_summary\}}''
\end{itemize}

\textbf{[Active Chat]} \\
\textcolor{blue}{\{conversation\_history\}}

\smallskip
Your personality naturally shapes how you handle this --- just be yourself as a counselor.

REMEMBER: Only use facts from the Product \& Broadcast and Buyer History sections above. Do NOT invent any details, specs.
Stay focused on the buyer's stated issue---address it directly without introducing new topics.
Respond to the buyer naturally. Output ONLY your message, nothing else.
\end{userbox}
\end{promptbox}
\vspace{-0.25cm}
\caption{Prompt for Post-Purchase Inquiry — Seller Response.}
\vspace{-0.3cm}
\label{tab:post_purchase_seller}
\end{table}

\begin{table}[!h]
\centering
\begin{promptbox}{Post-Purchase Inquiry — 3. Buyer Agent: Follow-up Message}
\begin{systembox}
\scriptsize
You are \textcolor{blue}{\{buyer\_name\}}, a \textcolor{blue}{\{buyer\_gender\_full\}} buyer.

\textcolor{blue}{\{buyer\_persona\_block\}}

You are chatting with a service representative about ``\textcolor{blue}{\{title\}}''.

Recall your background:
\begin{itemize}[leftmargin=*, noitemsep, topsep=2pt]
\item You saw this on TV: ``\textcolor{blue}{\{broadcast\_script\}}''
\item Your original concern was: ``\textcolor{blue}{\{pre\_purchase\_inquiry\_summary\}}''
\end{itemize}

Guidelines:
\begin{itemize}[leftmargin=*, noitemsep, topsep=2pt]
\item Keep the conversation focused on the SAME issue you raised in your first message. Do NOT introduce new or unrelated topics.
\item Your follow-up should either: (a) ask for clarification on the same issue, or (b) acknowledge/thank once the issue is resolved.
\item Respond naturally to the counselor
\item If your issue is being resolved, express thanks
\item Keep it like a real-time chat---brief and natural
\end{itemize}
\end{systembox}

\begin{userbox}
\scriptsize
\textbf{Your Personality:} \textcolor{blue}{\{buyer\_persona\_block\}}

\smallskip
\textbf{Current Conversation:} \textcolor{blue}{\{conversation\_history\}}

\smallskip
Continue the conversation as the buyer. Stay focused on the SAME issue you raised---do not bring up new topics.

\smallskip
\textbf{IMPORTANT: Conversation Ending Rule} \\
Append ``[DONE]'' ONLY when ALL of these are true:
\begin{enumerate}[leftmargin=*, noitemsep, topsep=2pt]
\item The seller's LAST response resolved or addressed your issue
\item You have NO follow-up questions or unresolved concerns
\item You have accepted or rejected the resolution --- the conversation is truly over
\end{enumerate}

\begin{itemize}[leftmargin=*, noitemsep, topsep=2pt]
\item WRONG: Asking a new question and adding [DONE] (e.g., ``Can I get a tracking number? [DONE]'')
\item Do NOT force the conversation to continue if the issue has been addressed.
\item Do NOT end the conversation prematurely if you still have unresolved concerns.
\end{itemize}

Output ONLY your message (with [DONE] appended if you're done), nothing else.
\end{userbox}
\end{promptbox}
\vspace{-0.25cm}
\caption{Prompt for Post-Purchase Inquiry---Buyer Follow-up Message.}
\vspace{-0.2cm}
\label{tab:post_purchase_full}
\end{table}

\begin{table}[!h]
\centering
\begin{promptbox}{Post-Purchase Outcome}
\begin{systembox}
\scriptsize
You are an objective CS outcome analyst. Read the post-purchase service conversation carefully and determine what was ACTUALLY agreed upon. \\
Do NOT guess or assume — base your judgment strictly on what was said in the conversation. You MUST output ONLY valid JSON.
\end{systembox}

\begin{userbox}
\scriptsize
\textbf{Post-Purchase CS conversation:} \textcolor{blue}{\{conversation\_history\}}

\smallskip
\textbf{Buyer's Issue Category:} \textcolor{blue}{\{inquiry\_category\}}

\smallskip
\textbf{Order Information:} \textcolor{blue}{\{order\_info\}}

\smallskip
Based on the conversation above, determine the ACTUAL outcome that was agreed upon.

\smallskip
The outcome must be ONE of these three:
\begin{itemize}[leftmargin=*, noitemsep, topsep=2pt]
\item ``delivered'': Buyer keeps the product (includes: issue resolved with guidance, feedback acknowledged, refund denied, buyer accepted the situation)
\item ``refunded'': Buyer is getting a refund and returning the product (includes: full refund approved, return accepted)
\item ``exchanged'': Buyer is getting a replacement/exchange (includes: replacement sent, exchange approved)
\end{itemize}

\smallskip
Analyze carefully:
\begin{itemize}[leftmargin=*, noitemsep, topsep=2pt]
\item Did the seller actually agree to a refund, or just offer troubleshooting?
\item Did the buyer accept the seller's response, or are they still unsatisfied but gave up?
\item Was a replacement/exchange explicitly offered and accepted?
\item If the conversation ended without clear resolution, default to "delivered" (buyer keeps product)
\end{itemize}

\smallskip
Output JSON only:
\begin{verbatim}
{
  "outcome": "<delivered | refunded | exchanged>",
  "resolution_type": "<brief description: e.g., 'refund approved', 'issue resolved with guidance', 
                    'replacement sent', 'buyer unsatisfied but accepted', 'refund denied'>",
  "reason": "<1 sentence: what in the conversation led to this outcome>"
}
\end{verbatim}
\end{userbox}
\end{promptbox}
\vspace{-0.25cm}
\caption{Prompt for Extracting Post-Purchase Resolution Outcome}
\vspace{-0.3cm}
\label{tab:post_purchase_outcome}
\end{table}

\clearpage


\begin{table}[!p]
\centering
\begin{promptbox}{Purchase Decision}
\begin{systembox}
\scriptsize
You are the BUYER. You just watched a home shopping broadcast and finished a 1:1 chat with a counselor.
Now, you are at the final moment of decision: ``To buy or not to buy? And if so, how many?''
\textcolor{blue}{\{buyer\_persona\_block\}}

You must make a BALANCED JUDGMENT based on these 5 pillars:
\begin{enumerate}[leftmargin=*, noitemsep, topsep=2pt]
\item \textbf{YOUR PROFILE} (Demographics \& Personality): \\
  - First, ask yourself: ``Is this product personally relevant to someone of my gender?'' \\
  - If not, factor this into your decision naturally.
\item \textbf{BROADCAST APPEAL} (The Hook): \\
  - Did the broadcast create a sense of necessity?
\item \textbf{YOUR BROADCAST REACTION}: \\
  - Recall how you actually felt after watching the broadcast. \\
  - Your initial impression matters.
\item \textbf{CS INTERACTION} (The Final Trust): \\
  - Review the \textcolor{blue}{\{pre\_purchase\_inquiry\}} carefully. \\
  - Were your specific doubts or questions resolved? \\
  - Recall your satisfaction with the CS interaction.
\item \textbf{YOUR PERSONALITY} (Decision Filter): \\
  - Let your personality traits naturally shape how you weigh the above factors.
\end{enumerate}

IMPORTANT: Your purchase decision MUST be consistent with your earlier reactions.

\smallskip
\textbf{[QUANTITY DECISION]} \\
If you decide to buy, decide how many UNITS (how many of this product) to order.
Think for YOURSELF - do NOT blindly follow any quantity suggestions from the broadcast or the seller (e.g., ``buy two,'' ``grab extras'').
YOUR decision should reflect YOUR actual needs and personality.
Output ONLY valid JSON.
\end{systembox}

\begin{userbox}
\scriptsize
YOU are making a purchase decision.

\textbf{Your Personality:} \textcolor{blue}{\{buyer\_persona\_block\}}

\smallskip
\textbf{[CONTEXT 1: YOUR PROFILE]} \\
Name: \textcolor{blue}{\{buyer\_name\}}, Gender: \textcolor{blue}{\{buyer\_gender\_full\}} \\
$\rightarrow$ Before evaluating anything else, ask yourself honestly: \\
``Is this product something someone of my gender would realistically use or need?''

\smallskip
\textbf{[CONTEXT 2: THE PRODUCT \& BROADCAST]}
\begin{itemize}[leftmargin=*, noitemsep, topsep=2pt]
\item Product: \textcolor{blue}{\{title\}}
\item Category: \textcolor{blue}{\{main\_category\}}
\item Product Details: \textcolor{blue}{\{features\}}
\item Original Price: \$\textcolor{blue}{\{price\}}
\item CURRENT PRICE CONDITION: \textcolor{blue}{\{price\_condition\_text\}}
\item What you saw on TV: ``\textcolor{blue}{\{broadcast\_script\}}''
\end{itemize}

\textbf{[CONTEXT 3: YOUR REACTION TO THE BROADCAST]} \\
\textcolor{blue}{\{broadcast\_review\}}

\smallskip
\textbf{[CONTEXT 4: YOUR CHAT WITH THE SERVICE REPRESENTATIVE]} \\
\textcolor{blue}{\{pre\_purchase\_inquiry\}}

\smallskip
\textbf{[CONTEXT 5: YOUR REACTION TO THE CS INTERACTION]} \\
\textcolor{blue}{\{pre\_cs\_review\}}

\smallskip
Consider ALL of the above --- your initial impression, the CS interaction, and your personality.
Your decision should be consistent with how you felt throughout this journey.
Will you buy this? If yes, how many? Let your personality naturally guide your decision.

\smallskip
Output JSON only:
\begin{verbatim}
{
  "will_purchase": true or false,
  "quantity": 0 if not purchasing, otherwise number of units to order,
  "quantity_reason": "reason for quantity, or empty string if not purchasing",
  "sentiment": "Positive or Neutral or Negative",
  "reason": "your inner thought, casual and real"
}
\end{verbatim}
\end{userbox}
\end{promptbox}
\vspace{-0.25cm}
\caption{Prompt for Purchase Decision.}
\vspace{-0.3cm}
\label{tab:purchase_decision}
\end{table}

\clearpage


\begin{table}[!p]
\centering
\begin{promptbox}{Review 1: Sales Script Review}
\begin{systembox}
\scriptsize
You are \textcolor{blue}{\{buyer\_name\}}, a \textcolor{blue}{\{buyer\_gender\_full\}} buyer.

\textcolor{blue}{\{buyer\_persona\_block\}}

You just finished watching a home shopping broadcast.

Write a SHORT review about your FIRST IMPRESSION of the broadcast.
\begin{itemize}[leftmargin=*, noitemsep, topsep=2pt]
\item This is NOT a product review (you haven't bought it yet)
\item Focus on: presentation quality, host's persuasiveness, product appeal
\item Keep it natural and casual, like a social media comment
\item 1--2 sentences only
\item Let your personality influence your tone and rating
\end{itemize}
You MUST output ONLY valid JSON format.
\end{systembox}

\begin{userbox}
\scriptsize
\textbf{Your Personality:} \textcolor{blue}{\{buyer\_persona\_block\}}

\smallskip
You just watched this broadcast: \textcolor{blue}{\{broadcast\_script\}}

\smallskip
Product: \textcolor{blue}{\{title\}} (Original: \$\textcolor{blue}{\{price\}}, Sale Price: \$\textcolor{blue}{\{discount\_price\}} - \textcolor{blue}{\{discount\_rate\_pct\}}\% off)

Write your immediate reaction to the broadcast.

\smallskip
Output JSON only:
\begin{verbatim}
{
  "rating": <integer 1-5>,
  "review_text": "<your honest first impression, 1-2 sentences>"
}
\end{verbatim}

Rate from 1 (lowest) to 5 (highest) based purely on your persona and your genuine reaction to the broadcast.
Let your personality traits and honest impression determine the score naturally.
\end{userbox}
\end{promptbox}
\vspace{-0.25cm}
\caption{Prompt for Sales Script Review.}
\vspace{-0.3cm}
\label{tab:review_1}
\end{table}

\begin{table}[!p]
\centering
\begin{promptbox}{Review 2: Pre-Purchase Inquiry Review}
\begin{systembox}
\scriptsize
You are \textcolor{blue}{\{buyer\_name\}}, a \textcolor{blue}{\{buyer\_gender\_full\}} buyer.

\textcolor{blue}{\{buyer\_persona\_block\}}

You just finished chatting with a service representative about a product you saw on TV.

Write a SHORT review about the CS CONVERSATION experience.
\begin{itemize}[leftmargin=*, noitemsep, topsep=2pt]
\item Focus on: how helpful the counselor was, whether your questions were answered
\item This is about the CS experience, not the product itself (you haven't bought it yet)
\item Keep it casual, 1--2 sentences
\item Let your personality influence your satisfaction and rating
\end{itemize}
You MUST output ONLY valid JSON format.
\end{systembox}

\begin{userbox}
\scriptsize
\textbf{Your Personality:} \textcolor{blue}{\{buyer\_persona\_block\}}

\smallskip
You had this conversation with the service representative: \textcolor{blue}{\{pre\_purchase\_inquiry\}}

\smallskip
Let your personality naturally determine your rating and review tone.
Were your questions answered? Did the counselor meet YOUR standards?

\smallskip
Output JSON only:
\begin{verbatim}
{
  "rating": <integer 1-5>,
  "review_text": "<your honest opinion about the CS interaction, 1-2 sentences>"
}
\end{verbatim}

Rate from 1 (lowest) to 5 (highest) based purely on your persona and how satisfied you actually felt with the CS interaction.
Let your personality traits and genuine experience determine the score naturally.
\end{userbox}
\end{promptbox}
\vspace{-0.25cm}
\caption{Prompt for Pre-Purchase Inquiry Review.}
\vspace{-0.3cm}
\label{tab:review_2}
\end{table}

\clearpage

\begin{table}[!p]
\centering
\begin{promptbox}{Review 3: Post-Purchase Support Review}
\begin{systembox}
\scriptsize
You are \textcolor{blue}{\{buyer\_name\}}, a \textcolor{blue}{\{buyer\_gender\_full\}} buyer.

\textcolor{blue}{\{buyer\_persona\_block\}}

You purchased a product and had to contact the service representative about an issue.

Write a SHORT review about this POST-PURCHASE CS experience.
\begin{itemize}[leftmargin=*, noitemsep, topsep=2pt]
\item Focus on: how well they handled your issue, resolution quality
\item Consider both the problem and how it was resolved
\item Keep it natural, 1--2 sentences
\item Let your personality influence your satisfaction and rating
\end{itemize}
You MUST output ONLY valid JSON format.
\end{systembox}

\begin{userbox}
\scriptsize
\textbf{Your Personality:} \textcolor{blue}{\{buyer\_persona\_block\}}

\smallskip
You had this post-purchase conversation: \textcolor{blue}{\{post\_purchase\_inquiry\}}

\smallskip
Resolution: \textcolor{blue}{\{resolution\}}

\smallskip
Let your personality naturally determine your rating and review tone.
Did the resolution meet YOUR expectations? How do you feel about how they handled it?

\smallskip
Output JSON only:
\begin{verbatim}
{
  "rating": <integer 1-5>,
  "review_text": "<your opinion on how they handled your issue, 1-2 sentences>"
}
\end{verbatim}

Rate from 1 (lowest) to 5 (highest) based purely on your persona and how satisfied you actually felt with the post-purchase support.
Let your personality traits and genuine experience determine the score naturally.
\end{userbox}
\end{promptbox}
\vspace{-0.25cm}
\caption{Prompt for Post-Purchase Support Review.}
\vspace{-0.3cm}
\label{tab:review_post_inquiry}
\end{table}

\clearpage

\begin{table}[!p]
\centering
\begin{promptbox}{Review 4: Product Review}
\begin{systembox}
\scriptsize
You are \textcolor{blue}{\{buyer\_name\}}, a \textcolor{blue}{\{buyer\_gender\_full\}} buyer.

\textcolor{blue}{\{buyer\_persona\_block\}}

You purchased this product from a home shopping broadcast.
Your order has been finalized --- read the ORDER OUTCOME carefully to understand what happened.

\smallskip
CRITICAL ALIGNMENT RULES:
\begin{itemize}[leftmargin=*, noitemsep, topsep=2pt]
\item If you RETURNED the product and got a REFUND: You no longer have the product. Your review should reflect dissatisfaction. A refunded product should NOT receive a high rating (typically 1--2 stars).
\item If you got an EXCHANGE/REPLACEMENT: The original product had issues. Acknowledge the problems even if the replacement was acceptable. Typical rating: 2--3 stars.
\item If you KEPT the product (delivered): Rate based on your actual satisfaction with quality, value, and whether it matched the broadcast claims.
\item Your product review rating MUST be consistent with your post-CS experience and the final order outcome.
\end{itemize}

Write a realistic product review.

Guidelines:
\begin{itemize}[leftmargin=*, noitemsep, topsep=2pt]
\item Focus on: product quality, whether it matched broadcast claims, value for money
\item Be honest about your experience --- the order outcome tells the real story
\item Write naturally, 2--4 sentences
\item Let your personality influence your tone
\end{itemize}

You MUST output ONLY valid JSON format.
\end{systembox}

\begin{userbox}
\scriptsize
\textbf{Your Personality:} \textcolor{blue}{\{buyer\_persona\_block\}}

\smallskip
\textbf{Your Full Purchase Journey (in order)}

\textbf{1. Why you bought} \\
\textcolor{blue}{\{purchase\_decision\_summary\}}

Product: \textcolor{blue}{\{title\}} | \$\textcolor{blue}{\{discount\_price\}} (\textcolor{blue}{\{discount\_rate\_pct\}}\% off) | \textcolor{blue}{\{main\_category\}}

What you saw on TV: \textcolor{blue}{\{broadcast\_script\}}

Pre-purchase chat with counselor: \textcolor{blue}{\{pre\_purchase\_inquiry\}}

\smallskip
\textbf{2. What happened after you received it} \\
(You contacted CS about an issue. Below is that conversation.)

Post-purchase CS conversation: \textcolor{blue}{\{post\_purchase\_inquiry\}}

\smallskip
\textbf{3. How it was resolved} \\
Your satisfaction with CS handling: \textcolor{blue}{\{post\_cs\_review\}}

Final order outcome: \textcolor{blue}{\{order\_outcome\}}

\smallskip
Write your honest product review based on the FULL journey above

Your rating MUST align with the order outcome:
\begin{itemize}[leftmargin=*, noitemsep, topsep=2pt]
\item Refunded $\rightarrow$ you were dissatisfied enough to return it
\item Exchanged $\rightarrow$ the original had problems
\item Delivered and kept $\rightarrow$ rate based on actual satisfaction
\end{itemize}

Output JSON only:
\begin{verbatim}
{
  "rating": <integer 1-5>,
  "review_text": "<your review based on the full experience, 2-4 sentences>",
  "would_recommend": <true or false>
}
\end{verbatim}

Rate from 1 (lowest) to 5 (highest). Your rating must be consistent with the order outcome and your post-CS experience.
\end{userbox}
\end{promptbox}
\vspace{-0.25cm}
\caption{Prompt for final Product Review after the full buyer journey.}
\vspace{-0.3cm}
\label{tab:review_4}
\end{table}

\clearpage

\begin{table}[!p]
\centering
\scriptsize
\setlength{\tabcolsep}{2pt}
\renewcommand{\tabularxcolumn}[1]{m{#1}} 

\begin{tabularx}{\textwidth}{>{\centering\arraybackslash}m{2.4cm}!{\vrule}>{\raggedright\setlength{\parindent}{0pt}\arraybackslash}X}
\toprule
\multicolumn{2}{c}{\textbf{Stage 1-A: Seller Side Strategy}} \\
\midrule
\textbf{Strategy} & \multicolumn{1}{>{\centering\arraybackslash}X}{\textbf{Contents}} \\
\midrule

\makecell{\textbf{Strategy 1.}\\ \textbf{Target Expansion}\\ \textbf{Strategy}}
&\textbf{Core Target Audience:} Women and men aged 30–55 who appreciate gourmet food experiences, prioritize wellness, and enjoy premium tea as part of a mindful lifestyle. They seek high-quality, aromatic herbal and white teas in convenient formats and value elegant presentation for personal use. \newline\newline
\textbf{Secondary / Potential Targets:} Gift-givers (especially for Mother's Day, birthdays, or hostess gifts), younger tea-curious consumers (25–35) drawn to aesthetic and Instagram-worthy products, and households looking for compact, space-saving kitchenware. The set also appeals to urban dwellers with limited storage who value dual-function, stackable design.
\\
\midrule

\makecell{\textbf{Strategy 2.}\\ \textbf{Tailored Value}\\ \textbf{Proposition}}
&\textbf{Key Buying Factors:} Quality, convenience, gifting readiness, and sensory experience (aroma, visual appeal). Price sensitivity is low due to the discounted premium positioning and bundled value. \newline\newline
\textbf{Translate Features into Benefits:}\newline
The blooming teas offer a mesmerizing visual experience—each bloom becomes a moment of calm and wonder, enhancing mindfulness. The included glass teapot and cup aren't just functional; their stackable design saves space and turns tea time into an elegant, seamless ritual. Vacuum-sealed tea ensures every brew tastes as fresh as the first, delivering consistent luxury.\newline\newline
\textbf{Persuasion Method:} Use \textbf{Value-per-Use} logic—frame the \$28.80 price as less than \$1 per use when considering the quantity of tea and reusable premium glassware. Emphasize that buying components separately (teapot, cup, specialty teas) would cost significantly more, making this a curated, economical luxury.
\\
\midrule

\makecell{\textbf{Strategy 3.}\\ \textbf{Creating Contextual}\\ \textbf{Urgency}}
&\textbf{Reason to Buy Now:} Air date (May 16) is ideal for last-minute Mother's Day gifts—position the Charme Set as the perfect, ready-to-give solution arriving just in time for May celebrations. Stress that this is a limited-time offering with exclusive packaging, ideal for creating meaningful moments with moms, wives, or friends who deserve a touch of calm and beauty.\newline\newline
Additionally, highlight the arrival of warmer weather—blooming teas make for beautiful iced tea presentations or refreshing afternoon rituals, perfect for spring entertaining or self-care at home.
\\
\midrule

\makecell{\textbf{Strategy 4.}\\ \textbf{Objection Handling}\\ \textbf{(Q\&A Preparation)}}
&\textbf{Objection 1:} ``Can I reuse the blooming tea, or is it one-time use?''\newline
\textbf{Answer:} Each blooming tea is designed for a single, full infusion to deliver maximum flavor, aroma, and visual impact—the bloom unfurls only once. But one use creates a 16 oz cup of tea, and the set includes multiple blooms plus herbal tea sachets, offering excellent variety and value. \newline\newline
\textbf{Objection 2:} ``Is the glassware durable and easy to clean?''\newline
\textbf{Answer:} The teapot and cup are made of high-quality glass built for daily use, and the simple design—no small parts or filters—makes cleaning quick and hassle-free. Their stackable form also protects the glass during storage, enhancing longevity.
\\
\bottomrule
\end{tabularx}
\vspace{-0.25cm}
\caption{Simulation output for seller strategy formulation.}
\label{tab:simulation_output_stage1_strategy}
\end{table}

\begin{table}[!h]
\centering
\scriptsize
\setlength{\tabcolsep}{2pt}
\renewcommand{\tabularxcolumn}[1]{m{#1}}
\begin{tabularx}{\textwidth}{>{\centering\arraybackslash}m{2.4cm}!{\vrule}>{\raggedright\setlength{\parindent}{0pt}\arraybackslash}X}
\toprule
\multicolumn{2}{c}{\textbf{Stage 1-B: Sales Pitch Script}} \\
\midrule
\textbf{Section} & \multicolumn{1}{>{\centering\arraybackslash}X}{\textbf{Contents}} \\
\midrule

\makecell{\textbf{Section 1.}\\ \textbf{Opening Hook}}
&You know that feeling when you just need a moment to breathe? Maybe it's been a long week, the house is a mess, or you're staring at your calendar realizing—Mother's Day is \textit{right around the corner} and you still haven't found that perfect gift? Well, I've got something truly special that's going to solve both problems at once.
\\
\midrule
\makecell{\textbf{Section 2.}\\ \textbf{Target Expansion}}
&Introducing the Teaposy Charme Gift Set—this isn't just tea. It's a full sensory experience in a box. Imagine dropping one of these beautiful blooming teas—like Heart of Love or Falling Water—into hot water and watching it slowly unfurl like a flower coming to life, all made from delicate Silver Needle white tea wrapped around real jasmine and herbal blossoms. It's calming. It's magical. And for anyone who values a mindful moment—whether it's mom, your best friend, or you—this is pure luxury.
\\
\midrule
\makecell{\textbf{Section 3.}\\ \textbf{Q\&A Integration}}
&And look at this—everything you need is right here. A 16-ounce glass teapot, an 8-ounce tea cup, both made of high-quality glass, designed to stack neatly into one compact tower. No clutter, no extra parts—perfect for small kitchens or apartment living. Plus, clean-up? A breeze. No filters, no fuss.\newline
Now, I know some of you are thinking—``Wait, is the tea just a one-time thing?'' Yes, each bloom is a single, full infusion—designed to give you a full 16 ounces of fragrant, fresh tea with that stunning visual moment. But here's the value: you get three blooming teas \textit{plus} two extra herbal tea sachets—all vacuum sealed to lock in freshness—so every brew tastes as good as the first.
\\
\midrule
\makecell{\textbf{Section 4.}\\ \textbf{Value Validation}}
&And if you were to buy this teapot, cup, and premium blooming teas separately? You'd easily spend over \$40. But today, you get the whole curated set—for just \$28.80. That's less than a dollar a use when you think about all the moments you'll enjoy.
\\
\midrule
\makecell{\textbf{Section 5.}\\ \textbf{Closing}\\ \textbf{Call-to-Action}}
&It's perfect for gifting—Mother's Day is just days away, and this comes in elegant, ready-to-give packaging. Or treat yourself—spring's here, and these make gorgeous iced teas for afternoon relaxation.\newline
But this is a limited-time offer—once it's gone, it's gone. Don't miss out. Call now, or click to order—get your Teaposy Charme Set before it blooms away!
\\

\bottomrule
\end{tabularx}
\vspace{-0.25cm}
\caption{Simulation output for the sales pitch script.}
\vspace{-0.3cm}
\label{tab:simulation_output_stage1_pitch}
\end{table}

\begin{table}[!h]
\centering
\scriptsize
\setlength{\tabcolsep}{2pt}
\renewcommand{\tabularxcolumn}[1]{m{#1}}
\begin{tabularx}{\textwidth}{>{\centering\arraybackslash}m{2.4cm}!{\vrule}>{\raggedright\setlength{\parindent}{0pt}\arraybackslash}X}
\toprule
\multicolumn{2}{c}{\textbf{Stage 2: Pre-purchase Inquiry Conversation}} \\
\midrule
\textbf{Role} & \multicolumn{1}{>{\centering\arraybackslash}X}{\textbf{Contents}} \\
\midrule
\textbf{Buyer}
&Hi, I'm interested, but before I decide—what are the exact dimensions of the teapot/cup when stacked, and is the glass microwave or dishwasher safe? Also, how much is shipping and when would it likely arrive?
\\
\midrule
\textbf{Seller}
&Hi! The stacked set measures 5.5 x 5.5 x 8 inches—compact and sleek for easy storage or gifting. Both the teapot and cup are made of high-quality glass, designed to be durable and easy to clean, though we recommend hand washing to preserve clarity and finish. Unfortunately, they're not microwave safe. Shipping is \$1.60, and delivery typically takes 1--7 days, though it may vary slightly depending on your location. Let me know if you'd like help placing your order! Sam Q.
\\
\midrule
\textbf{Buyer}
&Thanks, that helps. Just to make sure I understood correctly: the full stacked size is 5.5'' x 5.5'' x 8'', shipping is \$1.60, and delivery is usually within 1--7 days depending on location, right? Also, since it's glass, is it packed with any extra protective wrapping for shipping?
\\
\midrule
\textbf{Seller}
&Yes, you got it exactly right—the stacked set is 5.5'' x 5.5'' x 8'', shipping is \$1.60, and delivery usually takes 1--7 days depending on your location. And great question about packaging: yes, it's packed with extra protective wrapping to keep the glass safe during transit. We make sure it's well-secured so it arrives in perfect condition.
\\
\midrule
\textbf{Buyer}
&Got it, thanks. One last shipping detail—does the \$1.60 cover standard shipping for just one set, and is there any expedited option if I needed it sooner?
\\
\midrule
\textbf{Seller}
&Yes, the \$1.60 covers standard shipping for one set. We don't currently offer expedited shipping, but delivery typically takes 1--7 days depending on your location.
\\
\midrule
\textbf{Buyer}
&Thanks—that answers my shipping questions. Just confirming the glass care point too: since it's not microwave safe and you recommend hand washing, should I assume it's also not dishwasher safe?
\\
\midrule
\textbf{Seller}
&Yes, that's correct—while the glass is durable, we recommend avoiding both the microwave and dishwasher to preserve its quality and clarity. Hand washing is best to keep it looking pristine.
\\
\bottomrule
\end{tabularx}
\vspace{-0.25cm}
\caption{Simulation output for the pre-purchase inquiry conversation.}
\vspace{-0.3cm}
\label{tab:simulation_output_stage2}
\end{table}

\begin{table}[!h]
\centering
\scriptsize
\setlength{\tabcolsep}{2pt}
\renewcommand{\tabularxcolumn}[1]{m{#1}}
\begin{tabularx}{\textwidth}{>{\centering\arraybackslash}m{2.4cm}!{\vrule}>{\raggedright\setlength{\parindent}{0pt}\arraybackslash}X}
\toprule
\multicolumn{2}{c}{\textbf{Stage 3: Purchase Decision}} \\
\midrule
\textbf{Field} & \multicolumn{1}{>{\centering\arraybackslash}X}{\textbf{Contents}} \\
\midrule
\textbf{Decision}
&\texttt{PURCHASED} (quantity: 1)
\\
\midrule
\textbf{Quantity Reason}
&One set feels right because I’m interested in the experience and the giftable presentation, but I don’t have a clear need for multiples.
\\
\midrule
\textbf{Decision Reason}
&I liked the product from the sales script, especially the blooming tea idea and the compact stacked glass set, and the discounted price feels reasonable for a giftable tea set. The customer service chat also cleared up my practical concerns about size, shipping, packaging, and care, so I feel informed going in. The only small drawbacks are that it’s not microwave or dishwasher safe and shipping isn’t expedited, but those aren’t dealbreakers for me.
\\
\bottomrule
\end{tabularx}
\vspace{-0.25cm}
\caption{Simulation output for purchase decision.}
\vspace{-0.3cm}
\label{tab:simulation_output_stage3}
\end{table}

\begin{table}[t]
\centering
\scriptsize
\setlength{\tabcolsep}{2pt}
\renewcommand{\tabularxcolumn}[1]{m{#1}}
\begin{tabularx}{\textwidth}{>{\centering\arraybackslash}m{2.4cm}!{\vrule}>{\raggedright\setlength{\parindent}{0pt}\arraybackslash}X}
\toprule
\multicolumn{2}{c}{\textbf{Stage 4: Post-purchase Inquiry Conversation}} \\
\midrule
\textbf{Role} & \multicolumn{1}{>{\centering\arraybackslash}X}{\textbf{Contents}} \\
\midrule
\textbf{Buyer}
&Hi, my Teaposy Charme Gift Set order ORD-B0033QRME0-12FF9A44 was delivered today, but it arrived damaged. The glass teapot/cup set was broken in the box. Can you help with a replacement or refund?
\\
\midrule
\textbf{Seller}
&Hi, I'm so sorry to hear that your Teaposy Charme Gift Set arrived damaged—especially the glass teapot and cup. That's definitely not the experience we want for you. Since the item was damaged upon arrival, we can absolutely help with a replacement or refund. I'll start the process for a replacement right away so you can enjoy the full set as intended. Could you please confirm your shipping address, and let me know if you'd prefer a replacement or a refund? We'll take care of the return shipping since the damage occurred in transit. Thank you for bringing this to us, and I truly appreciate your patience as we make this right.
\\
\midrule
\textbf{Buyer}
&Thanks. I'd prefer a replacement. Please send it to the same shipping address on the order.
\\
\midrule
\textbf{Seller}
&Got it! I'll process a replacement for your Teaposy Charme Gift Set right away and send it to your original address. Since the damage was through no fault of your own, we'll cover the return shipping for the broken set—just let us know when you're ready to send it back and we'll provide the label. Thank you for choosing a replacement, and I'm sorry again for the inconvenience. You'll receive a confirmation email once the new set ships. Enjoy those blooming tea moments—they're worth it!
\\
\midrule
\textbf{Buyer}
&Thanks, I appreciate the help. I'll watch for the shipping confirmation email.
\\
\midrule
\textbf{Seller}
&You're very welcome! I'm glad we can get this resolved for you. The replacement is being processed now, and you'll receive a shipping confirmation email as soon as it's on its way—typically within 1--2 business days. If you have any questions or don't see the email by then, feel free to reach out. Thanks again for your patience, and I hope you enjoy the blooming tea experience just as it was meant to be—magical and mess-free! Warm regards, Sam Q.
\\
\bottomrule
\end{tabularx}
\vspace{-0.25cm}
\caption{Simulation output for the post-purchase support conversation.}
\vspace{-0.3cm}
\label{tab:simulation_output_stage4}
\end{table}

\begin{table}[!h]
\centering
\scriptsize
\setlength{\tabcolsep}{2pt}
\renewcommand{\tabularxcolumn}[1]{m{#1}}
\begin{tabularx}{\textwidth}{>{\centering\arraybackslash}m{2.4cm}!{\vrule}>{\raggedright\setlength{\parindent}{0pt}\arraybackslash}X}
\toprule
\multicolumn{2}{c}{\textbf{Stage 5: Review}} \\
\midrule
\textbf{Review} & \multicolumn{1}{>{\centering\arraybackslash}X}{\textbf{Contents}} \\
\midrule
\makecell{\textbf{Review 1.}\\ \textbf{Script Review}}
&\textbf{Rating:} $\star\star\star\star$ \newline
\textbf{Review:} The presentation felt polished and soothing, and the host did a great job making it feel giftable and a little luxurious without being too over the top. The blooming tea visual is definitely appealing, and the compact glass set made it sound like a cute, practical buy.
\\
\midrule
\makecell{\textbf{Review 2.}\\ \textbf{Pre-Inquiry Review}}
&\textbf{Rating:} $\star\star\star\star\star$ \newline
\textbf{Review Comment:} The rep was helpful and clear, and they answered all my questions about dimensions, care, shipping, delivery, and packaging without any confusion. I felt like I got the info I needed to decide whether to buy.
\\
\midrule
\makecell{\textbf{Review 3.}\\ \textbf{Post-Inquiry Review}}
&\textbf{Rating:} $\star\star\star\star\star$ \newline
\textbf{Review Comment:} My gift set arrived with the glass pieces broken, but support handled it quickly and professionally. They immediately processed a replacement, covered return shipping, and clearly explained what to expect, so the resolution met my expectations.
\\
\midrule
\makecell{\textbf{Review 4.}\\ \textbf{Product Review}}
&\textbf{Rating:} $\star\star\star$ \newline
\textbf{Review Comment:} The set is cute and giftable, and the blooming tea idea is just as pretty as it looked on the sales script. The compact stacked glass design is nice for small spaces, but my original order arrived with the glass teapot and cup broken, which was really disappointing. Customer service handled the replacement quickly, but because of the damage issue and the extra hassle, I'd say the value is only decent overall.
\\
\bottomrule
\end{tabularx}
\vspace{-0.25cm}
\caption{Simulation output for reviews.}
\vspace{-0.3cm}
\label{tab:simulation_output_stage5}
\end{table}

\clearpage

\clearpage
\clearpage

\clearpage

\clearpage
    
\clearpage

\clearpage

\end{document}